\documentclass{article}

\usepackage{amssymb}
\usepackage{amsthm}

\usepackage{booktabs, colortbl}       % professional-quality tables
\usepackage{microtype}      % microtypography

\usepackage{amsmath}
\DeclareMathOperator*{\argmax}{arg\,max}
\DeclareMathOperator*{\argmin}{arg\,min}

\usepackage[inline]{enumitem}
\usepackage{array} % for adjusting cell padding
\usepackage{mathtools} 
\usepackage{float}
\usepackage{soul}
\usepackage{multirow}
\usepackage{IEEEtrantools}
\usepackage{bbm}
\usepackage{graphicx}
\usepackage[export]{adjustbox}
\usepackage{tabulary}
\usepackage{tabularx}
\usepackage{natbib}

\newlist{mylist}{itemize}{1}
\setlist[mylist]{label=\textbf{ID:}}

\usepackage[pagebackref,breaklinks,colorlinks]{hyperref}

\usepackage[capitalize,noabbrev]{cleveref}
\newcommand{\supp}{\textcolor{black}{supplementary}}

\crefname{section}{Sec.}{Sec.}
\crefname{table}{Tab.}{Tab.}
\crefname{figure}{Fig.}{Fig.}
\crefname{equation}{Eq.}{Eq.}
\crefname{appendix}{Supp.}{Supp.}

\newenvironment{tight_itemize}{
\begin{itemize}[leftmargin=15pt,nosep]
  \setlength{\topsep}{0pt}
  \setlength{\itemsep}{0pt}
  \setlength{\parskip}{0pt}
  \setlength{\parsep}{0pt}
}{\end{itemize}}

\newenvironment{tight_enumerate}{
\begin{enumerate}[leftmargin=15pt]
  \setlength{\topsep}{0pt}
  \setlength{\itemsep}{0pt}
  \setlength{\parskip}{0pt}
  \setlength{\parsep}{0pt}
}{\end{enumerate}}

\newcommand{\para}[1]{\noindent \paragraph{#1}}
\newcommand{\Ours}{\texttt{LaGTran}}
\newcommand{\egoexo}{Ego2Exo}
\theoremstyle{plain}

\theoremstyle{definition}

\theoremstyle{remark}

\usepackage{amsmath}
\usepackage{amssymb}
\usepackage{mathtools}
\usepackage{amsthm}

\usepackage[capitalize,noabbrev]{cleveref}

\usepackage[accepted]{icml2024}
\icmltitlerunning{LaGTran: Language Guided Transfer Across Domains}

\usepackage{inconsolata}

\usepackage{caption}
\DeclareCaptionFont{xipt}{\fontsize{8}{10}\mdseries}
\usepackage[font=xipt]{caption}
\usepackage[font=xipt]{subcaption}

\begin{document}

\twocolumn[
\icmltitle{Tell, Don't Show!: Language Guidance Eases \\ Transfer Across Domains in Images and Videos}

% It is OKAY to include author information, even for blind
% submissions: the style file will automatically remove it for you
% unless you've provided the [accepted] option to the icml2024
% package.

% List of affiliations: The first argument should be a (short)
% identifier you will use later to specify author affiliations
% Academic affiliations should list Department, University, City, Region, Country
% Industry affiliations should list Company, City, Region, Country

% You can specify symbols, otherwise they are numbered in order.
% Ideally, you should not use this facility. Affiliations will be numbered
% in order of appearance and this is the preferred way.
\icmlsetsymbol{equal}{*}

\begin{icmlauthorlist}
\icmlauthor{Tarun Kalluri}{yyy}
\icmlauthor{Bodhisattwa Prasad Majumder}{comp}
\icmlauthor{Manmohan Chandraker}{yyy}
\\
\url{https://tarun005.github.io/lagtran/}
\end{icmlauthorlist}

\icmlaffiliation{yyy}{UC San Diego}
\icmlaffiliation{comp}{Allen Institute for AI}

\icmlcorrespondingauthor{TK}{sskallur@ucsd.edu}

% You may provide any keywords that you
% find helpful for describing your paper; these are used to populate
% the "keywords" metadata in the PDF but will not be shown in the document
\icmlkeywords{Domain Adaptation, Transfer Learning, Multimodal Learning, Vision-language Learning}

\vskip 0.3in
]

% this must go after the closing bracket ] following \twocolumn[ ...

% This command actually creates the footnote in the first column
% listing the affiliations and the copyright notice.
% The command takes one argument, which is text to display at the start of the footnote.
% The \icmlEqualContribution command is standard text for equal contribution.
% Remove it (just {}) if you do not need this facility.

%\printAffiliationsAndNotice{}  % leave blank if no need to mention equal contribution
\printAffiliationsAndNotice{\icmlEqualContribution} % otherwise use the standard text.

\begin{abstract}
    We introduce \Ours{}, a novel framework that utilizes text supervision to guide robust transfer of discriminative knowledge from labeled source to unlabeled target data with domain gaps. While unsupervised adaptation methods have been established to address this problem, they show limitations in handling challenging domain shifts due to their exclusive operation within the pixel-space. Motivated by our observation that semantically richer text modality has more favorable transfer properties, we devise a transfer mechanism to use a source-trained text-classifier to generate predictions on the target text descriptions, and utilize these predictions as supervision for the corresponding images. Our approach driven by language guidance is surprisingly easy and simple, yet significantly outperforms all prior approaches on challenging datasets like GeoNet and DomainNet, validating its extreme effectiveness. To further extend the scope of our study beyond images, we introduce a new benchmark called \egoexo{} to study ego-exo transfer in videos and find that our language-aided approach \Ours{} yields significant gains in this highly challenging and non-trivial transfer setting. Code, models and proposed datasets are publicly available at \url{https://tarun005.github.io/lagtran/}. 
\end{abstract}

\section{Introduction}
\label{sec:intro}

Despite great strides in the performance in several applications of computer vision recent years, achieving robustness to distribution shifts at test-time still remains a challenge. 
In particular, a fundamental need to improve generalization to domains without manual supervision arises due to the cost and scarcity of acquiring labeled images. 
A dominant paradigm to address this limitation has been unsupervised domain adaptation (UDA), which uses labels from a related source domain along with distribution alignment techniques to bridge the domain gap~\cite{DANN, CDAN, saito2018maximum, xu2019larger, sharma2021instance, wei2021toalign, Chen_2022_CVPR, zhu2023patch}. Despite their noted success, their limitations in addressing challenging transfer beyond regular domain shifts~\cite{saenko2010adapting, venkateswara2017deep, peng2017visda} is recently highlighted~\cite{prabhu2022can,kalluri2023geonet}. 
We posit that a part of this limitation potentially stems from their dependence on pixel-level data alone to bridge domain gaps, as accurately characterizing shifts and devising bridging strategies solely based on images becomes challenging beyond standard domain shift scenarios. 
% that a fundamental limitation in current adaptation algorithms lies in their dependence on pixel-level data alone to bridge domain gaps, as accurately identifying specific types of shifts and devising effective bridging strategies based solely on images becomes challenging beyond standard domain shift scenarios. 
% While a potential alternative lies in leveraging large-scale pre-trained vision-language models, 

\begin{figure}
    \centering
    \resizebox{0.48\textwidth}{!}{%
    \includegraphics{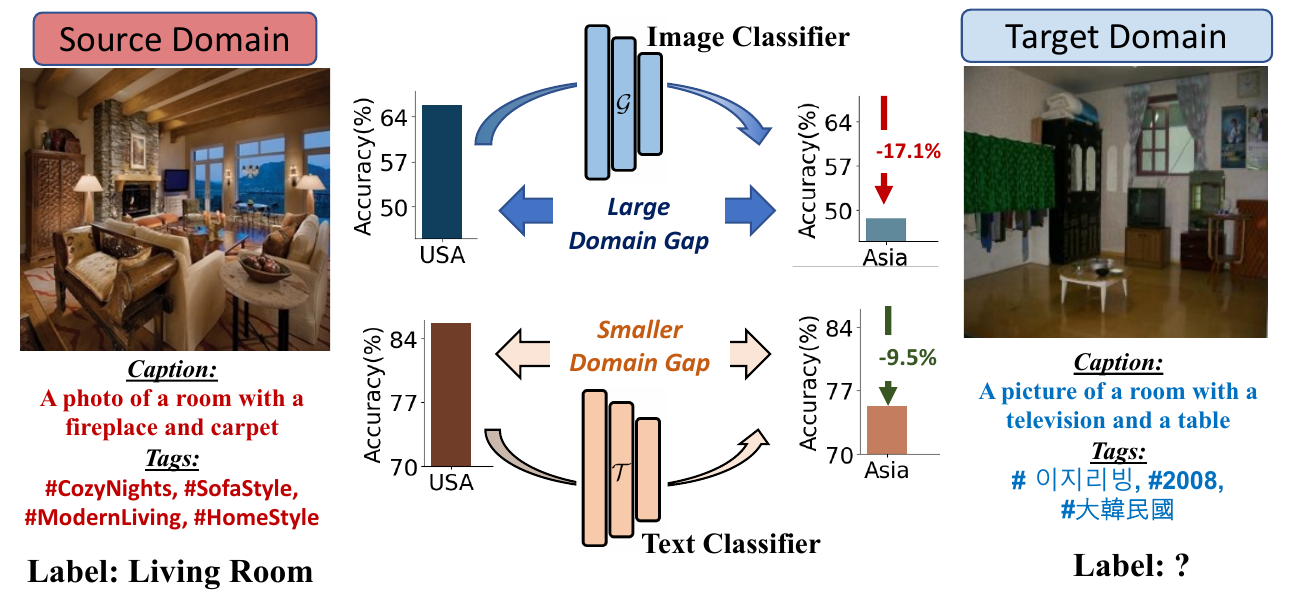}
    }
    \captionsetup{width=0.48\textwidth}
    \caption{{\bf A summary of our insights for \Ours{}}: 
    In a domain transfer setting with labeled source and unlabeled target domain data, we observe significantly more drop incurred while transferring an image-classifier trained on source images to target ($17.1\%$), compared to a text-classifier trained on corresponding text descriptions of source images ($9.5\%$). We use this insight to build a simple framework called \Ours{} that leverages these text descriptions easily available in both domains to improve transfer in images and videos.
    } 
    \label{fig:intro_pic}
    % \vspace{-9pt}
\end{figure}

In contrast, we propose an alternative approach to ease transfer across such challenging shifts by instead leveraging ubiquitously available language guidance during training. Our framework, called \textbf{\Ours{}} for \textbf{La}nguage \textbf{G}uided \textbf{Tr}ansfer \textbf{A}cross Domai\textbf{n}s, is surprisingly simple to implement, yet shows extreme effectiveness and competence in handling transfer across challenging domain shifts in images and videos compared to any image-based adaptation method. 
%
% To address this limitation, we propose a simple and highly effective alternative framework called \Ours{}, \textbf{La}nguage \textbf{G}uided \textbf{Tr}ansfer \textbf{A}cross Domai\textbf{n}s, that directly leverages ubiquitously available language supervision during training to improve cross-domain robustness. 
Our key insight lies in observing that text guidance, which is readily available in the form of metadata for internet-sourced datasets or easily generated with emerging image captioning models, requires no human annotation while offering a more suitable avenue in transferring discriminative knowledge even across challenging domain shifts. 

We further illustrate this property in \cref{fig:intro_pic}, where we examine the transferability of image and text classifiers trained using image or text supervision respectively between USA and Asia domains from the GeoNet dataset~\cite{kalluri2023geonet}. We observe significantly less drop ($9.5\%$) when applying a text classifier trained on the source text to target text, compared to $17.1\%$ drop incurred when transferring an image classifier to target images.
% ~\footnote{These observations hold for other domains with different text sources as well, as shown in the \supp{}.}. 
As text operates in a significantly lower-dimensional space, language modality naturally has lesser domain gaps as opposed to images or videos. 
Furthermore, text descriptions often contain valuable attributes and identifiers that enhance the ability to accurately recognize images in a standard classification setting, suggesting more favorable domain robustness and discriminative properties of language descriptions compared to images. 
% , and possess better discriminative properties in classifying images. 

In the current work, we incorporate these observations to improve transfer in a scenario where the source domain has text descriptions accessible along with the labels, but the target domain only has text descriptions corresponding to the images. Accordingly, we first train a text classifier using the source domain language descriptions and labels and transfer this classifier to assign pseudo-labels to the target text descriptions, which, from \cref{fig:intro_pic}, would yield more robust pseudo-labels compared to the common image-based transfer~\cite{liu2021cycle, sun2022safe, kumar2020understanding}.
% , eliminating the need for advanced refinement modules~\cite{sun2022safe} or curriculum training~\cite{kumar2020understanding}. 
% Furthermore, the already low domain gaps in language without 
We, therefore, directly use these pseudo-labels as supervision for the unlabeled target images to train an image classifier jointly with source labels. This simple technique, free of any complicated adaptation mechanisms, shows remarkably strong performance
% on challenging datasets like GeoNet~\cite{kalluri2023geonet} and DomainNet~\cite{peng2019moment}, 
surpassing competitive baselines and prior UDA methods. 

To further demonstrate the broad usefulness of \Ours{} beyond images, we introduce and study a novel benchmark for transfer learning in videos called \egoexo{}, which focuses on the previously under-explored challenge of transferring action recognition between ego (first-person) and exo (third-person) perspectives in videos~\cite{li2021ego, ohkawa2023exo2egodvc, quattrocchi2023synchronization, xue2024learning} from a transfer learning standpoint. 
% While prior works design algorithms 
% addressing transfer between the views in supervised or self-supervised scenarios~\cite{}, a comprehensive dataset to study is lacking. 
We curate \egoexo{} benchmark using cooking videos from the Ego-Exo4D dataset utilizing key step annotations to assign action labels and atomic action descriptions for textual guidance. Our language-aided transfer shows remarkable utility in this challenging setting, significantly outperforming prior Video-UDA methods~\cite{chen2019temporal, wei2022unsupervised}.

In summary, our contributions are three-fold.
\begin{tight_itemize}
    \item A novel framework \Ours{} highlighting the feasibility of incorporating various forms of readily available text supervision in enhancing transfer across domain shifts (\cref{subsec:ours_details}).

    \item A new dataset \egoexo{} to study the problem of cross-view transfer in videos with fine-grained labels covering a diverse pool of actions and free-form text descriptions providing language guidance (\cref{subsec:videos}). 

    \item Demonstration of the competence of \Ours{} across a variety of domain shifts, with non-trivial gains over UDA methods on challenging datasets like GeoNet (+10\%), DomainNet (+3\%) and the proposed \egoexo{} (+4\%) datasets (\cref{sec:expmnts}). 
\end{tight_itemize}

\section{Related Work}
\label{sec:related_work}

\vspace{-12pt}
\para{Domain robustness in computer vision.}
% Our work studies the setting where it is necessary to improve accuracy on an unlabeled target domain by leveraging labels from a different source domain. 
%
A suite of methods have been proposed to improve accuracy on an unlabeled target domain by leveraging labels from a different source domain using unsupervised adaptation~\cite{ben2006analysis, ben2010theory, DANN, CDAN}, where prior works propose various domain alignment strategies including MMD-based~\cite{tan2020class, long2017deep, sun2016deep, kang2019contrastive}, adversarial~\cite{bousmalis2016domain, tzeng2017adversarial, saito2017adversarial, chen2019progressive, tzeng2015simultaneous, wei2021toalign}, class-specific adaptation~\cite{pei2018multi, saito2018maximum, luo2019taking, xie2018learning, kumar2018co, gu2020spherical}, clustering~\cite{deng2019cluster, park2020joint, li2021semantic, kalluri2022cluster} instance-specific adaptation~\cite{sharma2021instance, kalluri2022memsac, wang2022cross}, self-training~\cite{french2017self, liu2021cycle, sun2022safe, prabhu2022can} and more recently transformers~\cite{xu2021cdtrans} or patch-based mechanisms~\cite{zhu2023patch}. Similar ideas have also been explored in video domain adaptation~\cite{choi2020unsupervised}, with extensions to incorporate temporal alignment ~\cite{chen2019progressive, wei2022unsupervised, sahoo2021contrast, dasgupta2022overcoming}. However, all these uda methods predominantly operate in the pixel-space, and often fall short in bridging more complicated forms of domain shift in challenging transfer settings~\cite{kalluri2023geonet, prabhu2022can}. While some contemporary efforts utilize pre-trained CLIP models~\cite{lai2023padclip, lai2024empowering, zhang2023domain} or LLMs~\cite{chen2024large} for domain alignment, we show that per sample natural language guidance can be equally effective. 
Based on this, our work introduces a new paradigm to study robustness, where we build a simple framework using rich textual descriptions to overcome large domain gaps in image and video recognition tasks. 
%

% \vspace{-6pt}
\para{Language supervision in computer vision.}

% In contrast to models trained only on images using supervised or unsupervised datasets, recent works highlight the effectiveness of including 

The recent proliferation of internet-sourced datasets highlights the ready availability of natural language supervision without the need for any labeling or annotation efforts in images~\cite{thomee2016yfcc100m, changpinyo2021conceptual, schuhmann2022laion, mahajan2018exploring, desai2021redcaps} and videos~\cite{miech2019howto100m, bain2021frozen, grauman2022ego4d, grauman2023ego}. This availability of language supervision has been effectively utilized to learn scalable weakly supervised models~\cite{mahajan2018exploring, singh2022revisiting}, robust vision-language representations~\cite{radford2021learning, jia2021scaling, pham2023combined, desai2021virtex, sariyildiz2020learning, lin2022egocentric, zhao2023learning, goyal2022vision}, text-conditioned generative models~\cite{rombach2022high, ramesh2021zero, saharia2022photorealistic} and improving sampling techniques for self-supervised learning~\cite{el2023learning}.
% , where the use of language supervision has been shown to confer additional grounding capabilities and improved generalization in visual representations. 
%
Even in the absence of associated language supervision, recent innovations showed the potential of generating correlated descriptions for images using image-to-text or image captioning models~\cite{li2023blip2, liu2023visual, achiam2023gpt}. 
%
% Although pre-training using language supervision shows strong zeroshot inference capability, substantial improvements are possible 
%
Despite this ubiquity and proven effectiveness of language supervision for vision tasks, little attention has been directed at leveraging their utility in improving transfer learning across domains. In this work, we use language guidance to develop a straightforward mechanism to improve image and video classification on domains without manual supervision.
% In this work, we seek to address this gap by first demonstrating the favorable domain robustness properties inherent in language as opposed to images, and using these insights in developing a straightforward mechanism to improve image and video classification on domains without manual supervision.

\para{Domain robustness using language supervision.} Recent emergence of large-scale pre-trained vision-language foundational models such as CLIP~\cite{radford2021learning} enabled strong zero-shot generalization across diverse domains and tasks~\cite{devillers2021does}. However, the zeroshot inference using frozen pre-trained models still fall short of supervised fine-tuning~\cite{radford2021learning, pham2023combined, andreassen2021evolution}, which in-turn suffers from poor generalization to distributions outside the fine-tuning data~\cite{kumar2022fine, wortsman2022robust}. Prior works explored robust fine-tuning of zero-shot models, but do not leverage target domain data~\cite{udandarao2023sus} or language supervision~\cite{wortsman2022robust} during fine-tuning. While recent works incorporate language guidance for domain generalization~\cite{dunlap2023using, wang2024landa, liu2023tdg, huang2023sentence, min2022grounding, gokhale2021attribute}, they mostly rely on domain or class descriptors and do not leverage semantically richer free form text supervision from target images during transfer. One work which is closest to ours is \cite{goyal2023finetune}, which uses label names as text descriptions while we use free-form captions for images or atomic annotations for actions.  In video recognition literature, prior works seek to align ego and exo views using language pre-training~\cite{xu2024retrieval, huang2024egoexolearn}. 
In contrast to these efforts, we show that incorporating language aided transfer through diverse supervision yields a remarkably effective framework for improving domain robustness in both images and videos. 

\section{Method Details}
\label{sec:method}

% We provide an overview of \Ours{} in \cref{fig:}. In \Ours{}, we first train a discriminative 

\subsection{Problem Description and Background} 
\label{subsec:background}

%%%%%%%%%%%%%%%%%%%%%%%%%%%%%%%%%%%%%%%%%%%
%%%%%%%%%%%%%%%%%%%%%%%%%%%%%%%%%%%%%%%%%%
\begin{figure}[!tbp]
  \begin{minipage}{0.47\textwidth}
    \centering
    \includegraphics[width=\linewidth]{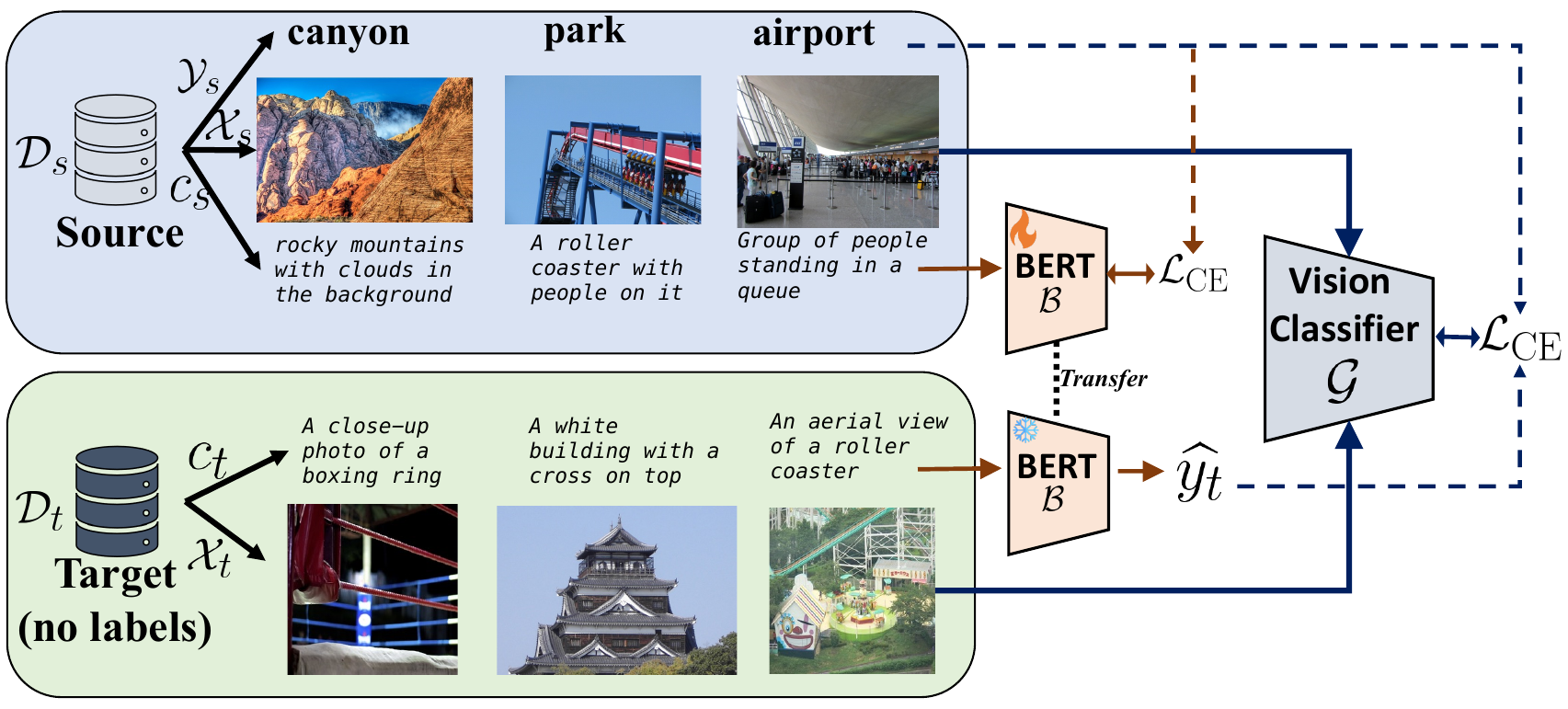}
    \captionsetup{width=0.98\textwidth}
    \caption{{\bf An overview of training using \Ours{}:} We operate in a setting where the labeled source domain and unlabeled target domain data possess cheaply available or easily generated language descriptions for each image. \Ours{} proceeds by first training a \textcolor{orange}{BERT-classifier} $\mathcal{B}$ using source captions and labels (\cref{eq:text_Classify}), and using the trained model to generate pseudo-labels $\hat{y}_t$ for the target captions and corresponding images (\cref{eq:pseudo}). We then use this generated supervision along with source domain data in jointly training a \textcolor{darkgray}{\textbf{Vision classifier}} $\mathcal{G}$ for image or video classification (\cref{eq:joint_training}).}
    \label{fig:method_figure}
  \end{minipage}%
  % \vspace{-12pt}
\end{figure}
%%%%%%%%%%%%%%%%%%%%%%%%%%%%%%%%%%%%%%%%%%
%%%%%%%%%%%%%%%%%%%%%%%%%%%%%%%%%%%%%%%%%%%

% we study the problem of unsupervised transfer across domainsm where we are need to improve performance on certain target domains with no manual supervision using labels from a different source domain data.  

\noindent We consider the setting of unsupervised cross-domain transfer, with access to labeled data from a source domain $\mathcal{D}_s:\{X_s^i,y_s^i\}_{i=1}^{N_s}$ along with unlabeled data from a target domain $\mathcal{D}_t:\{X_t^i\}_{i=1}^{N_t}$, where $X_s {\sim} P_s$, $X_t {\sim} P_t$, $N_s$ and $N_t$ are the number of samples in source and target domains, and the covariate shift assumption means marginal distributions $P_s \neq P_t$~\citep{ben2006analysis, ben2010theory}.
However, different from prior works, we additionally assume access to natural language descriptions, denoted by $c_i$, corresponding to each image or video input in both source and target domains during training. Consequently, we denote the labeled source domain with $\mathcal{D}_s:\{X_s^i,y_s^i,c_s^i\}_{i=1}^{N_s}$ and the unlabeled target domain with $\mathcal{D}_t:\{X_t^i,c_t^i\}_{i=1}^{N_t}$. These text descriptions are readily available through associated metadata in web-collected images~\cite{mahajan2018exploring}, or can be effortlessly generated with state-of-the-art image-to-text models~\cite{li2023blip2}. 
In \cref{sec:expmnts}, we show robust performance using text descriptions derived from a variety of sources, including: image metadata (e.g., alt-text, hashtags) for web-sourced images, state-of-the-art image captioners for manually curated datasets, as well as action descriptions or narrations in videos.
Note that our setting requires language descriptions $c_i$ only during training and not during inference or deployment, and therefore incurs no speed or memory overhead at test-time when compared with prior approaches. 
%
% While traditional UDA methods that only work in pixel-space fail to address several challenging domain shift setting, often resulting in worse performance than a source-only baseline~\cite{kalluri2023geonet}, we show that our approach efficiently utilizes prevalent multimodal supervision and achieves significant gains over all prior methods and baselines (\cref{sec:expmnts}). 

\subsection{\Ours{} for Cross-Domain Transfer}
\label{subsec:ours_details}

% \vspace{-12pt}

\para{Overview.}
The training pipeline used in \Ours{} for cross-domain transfer is summarized in \cref{fig:method_figure}, where we 
% Our proposed approach is derived from the insight that language descriptions typically have less severe domain gaps and more discriminative information than corresponding images or videos, which we efficiently exploit in a cross-modal supervision transfer framework. In particular, 
first train a BERT sentence classifier using the (text, label) pairs from the source domain dataset, and utilize this trained classifier to infer predictions on all the descriptions from the target domain. We then use these predictions as pseudo-labels for the target images, and train a joint vision classifier along with the labeled source domain images. 
%

% \vspace{-12pt}
\para{Training the text classifier.}

% As shown in \cref{fig:intro_pic}, the target prediction accuracy of a network trained on text descriptions is much better than one trained on images or videos, which validates our hypothesis that the cross-domain gaps are less severe in text-space than pixel-space.
% In addition to the insights presented in \cref{fig:intro_pic}, 
% As shown in \cref{fig:introb}, a model trained to classify source text captions transfers , which validates our hypothesis that the cross-domain gaps are less severe in text-space than pixel-space. 

% which we use in creating synthetic supervision for the unlabeled target images. 

We use the supervised text-label pairs from the source domain $(c_i^s,y_i^s)$ and train a BERT~\cite{devlin2019bert} sentence classifier $\mathcal{B}$ to predict the category label from an input text description, using the training objective
\begin{equation}
    \phi^* = \argmin_{\phi} \mathbb{E}_{(c_i,y_i) \sim \mathcal{D}_s} \mathcal{L}_{\text{CE}} (\mathcal{B}(c_i;\phi), y_i) ,
    \label{eq:text_Classify}
\end{equation}
where $\phi$ denotes the parameters of the BERT classifier and $\mathcal{L}_\text{CE}$ is the supervised cross-entropy loss. We adopt a pre-trained Distill-BERT~\cite{sanh2019distilbert} model from HuggingFace as the sentence classifier $\mathcal{B}(;\phi)$, and fine-tune it on the source domain data. We observed sub-optimal performance using other pre-trained backbones such as T5~\cite{raffel2020exploring} or GPT-2~\cite{radford2019language} (\cref{tab:text_classifier}).
% or text encoder in CLIP~\cite{radford2021learning} . 
% and fine-tune it for five epochs over the source domain data using AdamW optimizer with a learning rate of 5e-5 and cosine decay over the training schedule.  
% We explore alternative options for the sentence classifier network $\mathcal{B}$ in , and observed 
% 
Across all datasets and experiment settings used in this paper, we feed the raw text descriptions directly into the sentence classifier network without any preprocessing or manual curation. 
We observed remarkable robustness of the trained classifier in handling several challenges posed by unfiltered text, including their variable lengths across images, language barriers prevalent in geographically diverse data, unrelated tags and descriptions commonly found in web-sourced images or potentially imperfect captions from state-of-the-art captioning models. 
% \Ours{} efficiently handles diverse forms of available text supervision. 

\begin{figure}
  \centering
  \begin{minipage}{.49\textwidth}
    \centering
    {%
      \begin{minipage}{.48\textwidth}
        \includegraphics[width=\linewidth]{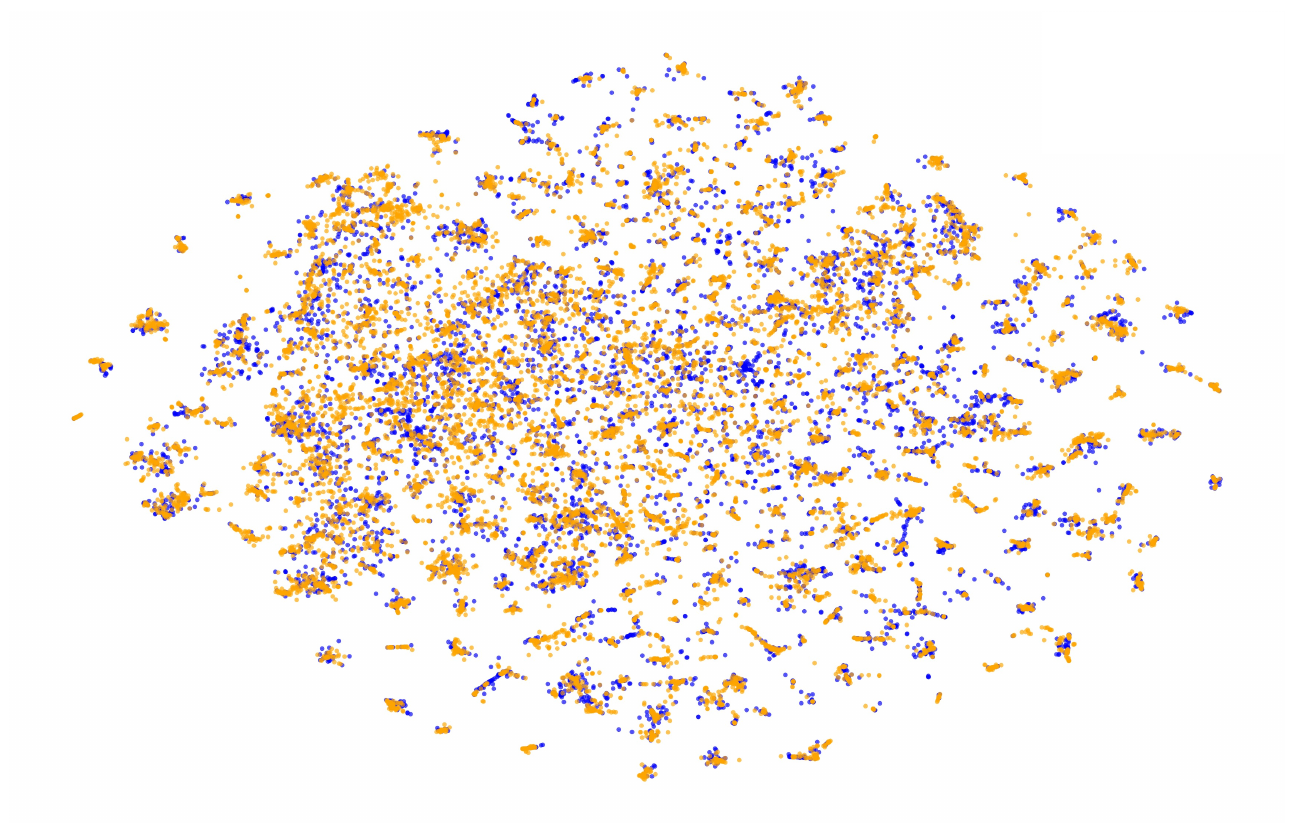}
        \subcaption{}
        \label{fig:geoimnet-image}
      \end{minipage}%
      \hfill
      \begin{minipage}{.48\textwidth}
        \includegraphics[width=\linewidth]{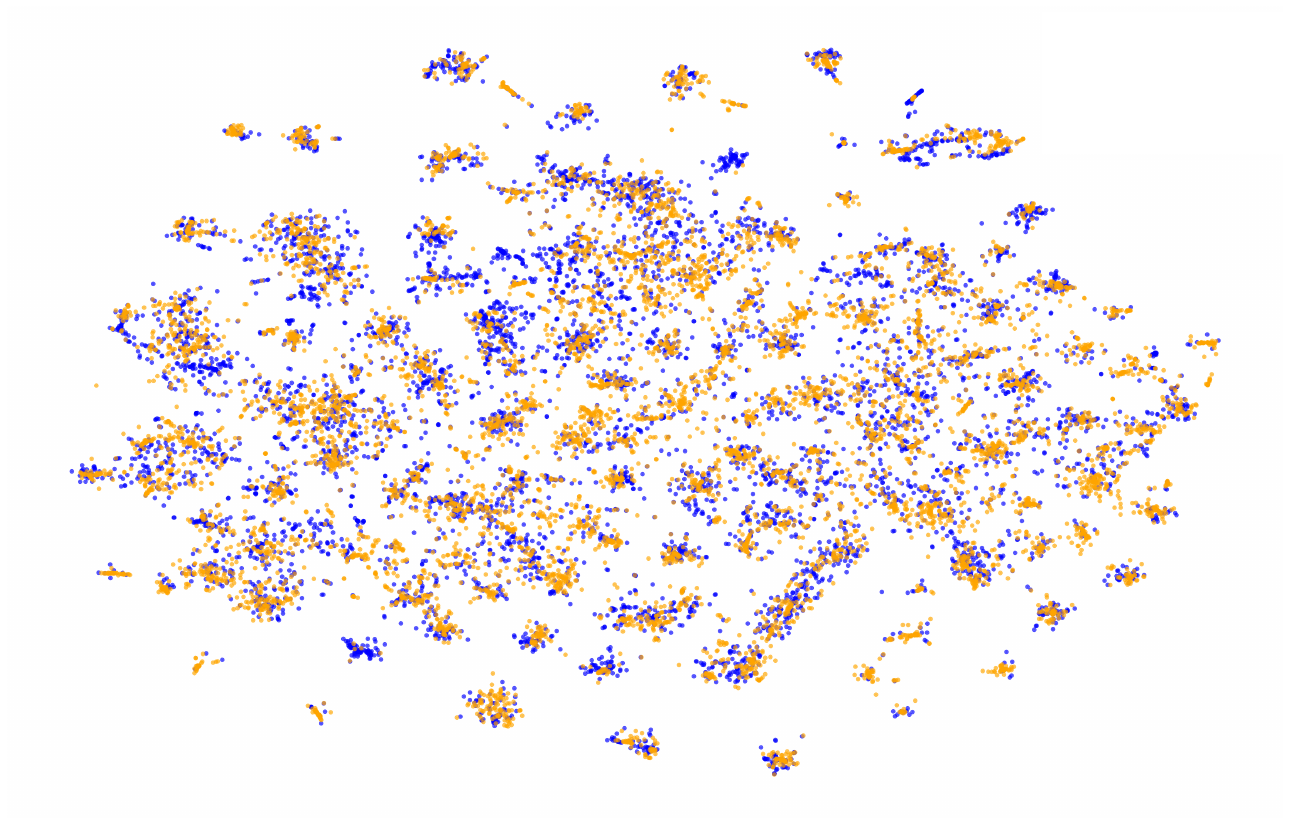}
        % \subcaption{GeoPlaces-image}
        \subcaption{}
        \label{fig:geoplaces-image}
      \end{minipage}%
    }
    
    {%
      \begin{minipage}{.48\textwidth}
        \includegraphics[width=\linewidth]{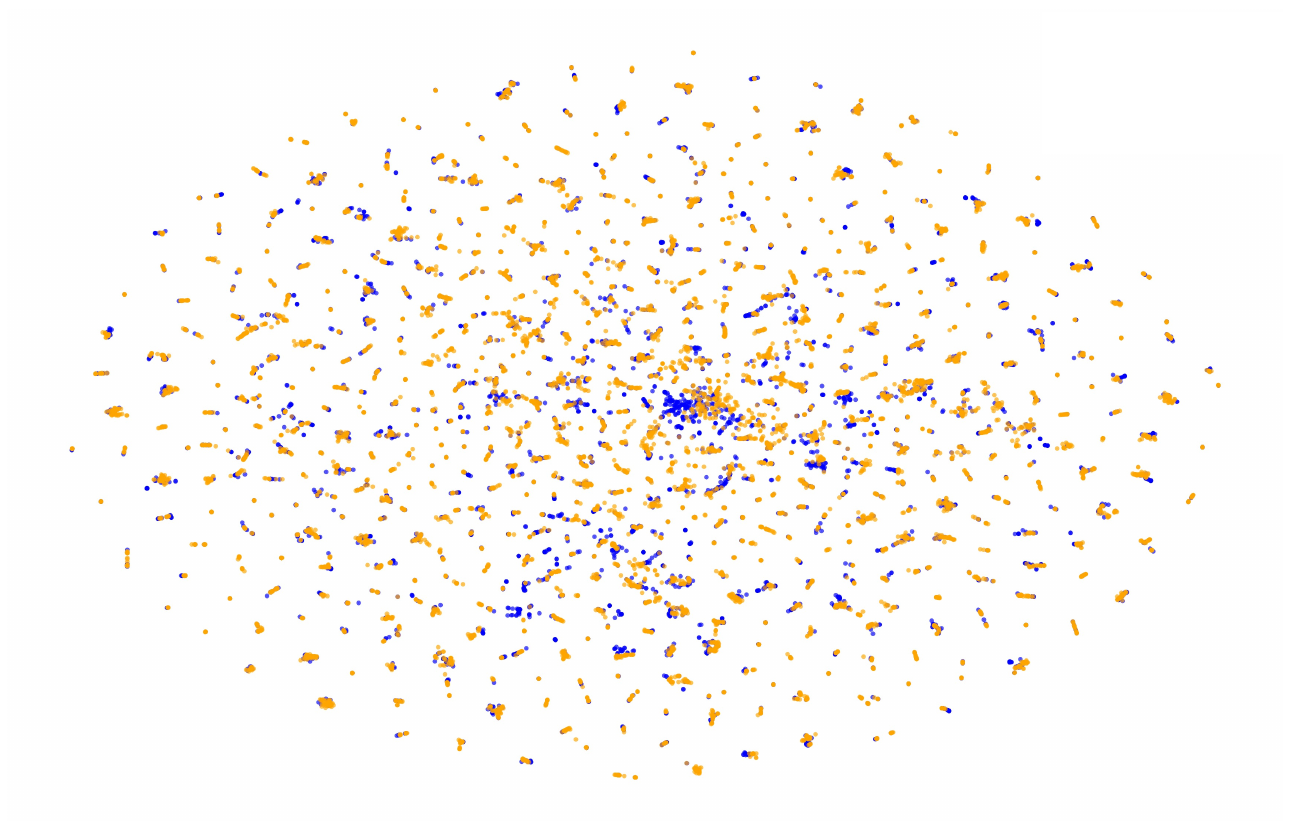}
        \subcaption{}
        \label{fig:geoimnet-text}
      \end{minipage}%
      \hfill
      \begin{minipage}{.48\textwidth}
        \includegraphics[width=\linewidth]{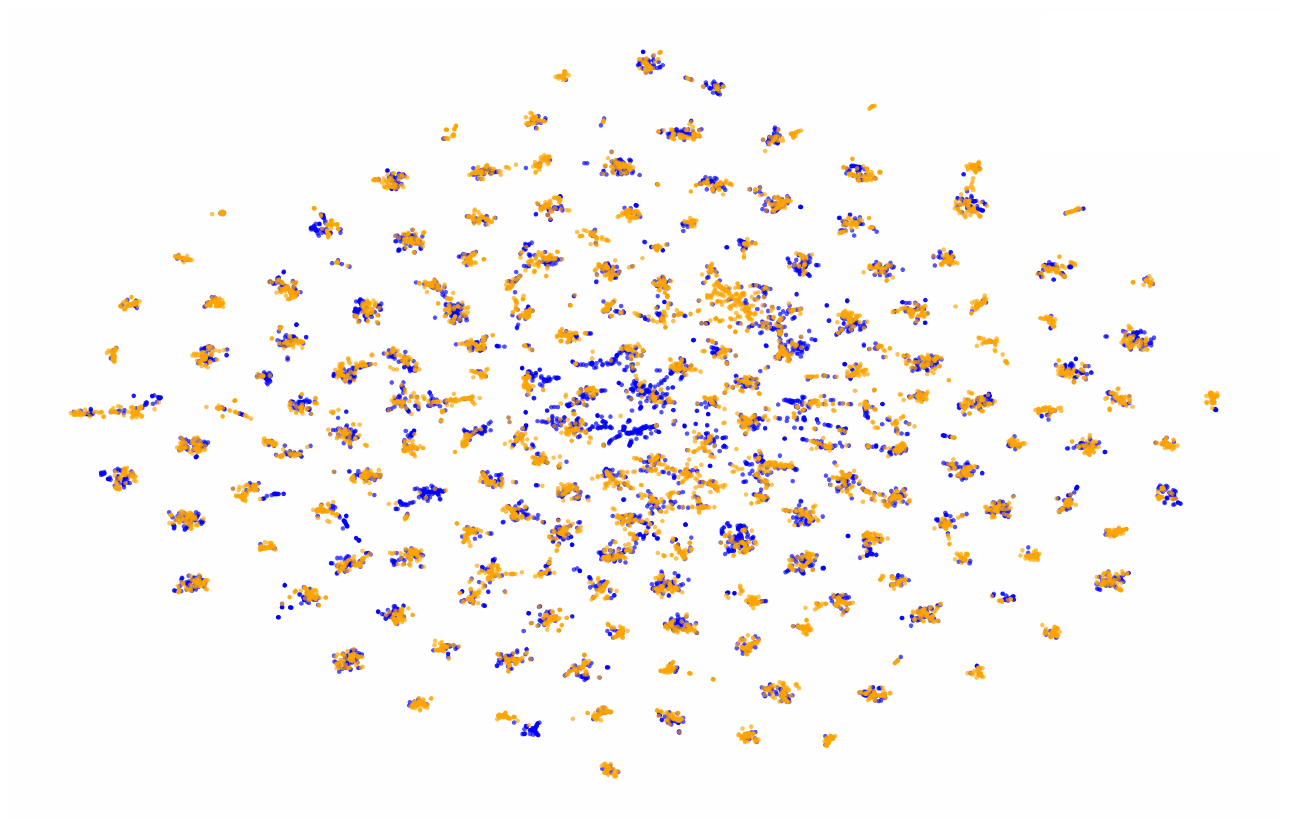}
        % \subcaption{GeoPlaces-text}
        \subcaption{}
        \label{fig:geoplaces-text}
      \end{minipage}%
    }
    \captionsetup{width=0.98\textwidth}
    \caption{{\bf tSNE visualization of cross-domain features on GeoNet.} We show improved domain-alignment with better class-separation in source and target when extracting features from a text-classifier (\crefrange{fig:geoimnet-text}{fig:geoplaces-text}) compared to features from image-classifier (\crefrange{fig:geoimnet-image}{fig:geoplaces-image}) highlighting better transferability through text modality.
    (\textcolor{orange}{Source in orange} and \textcolor{blue}{target in blue}).}
    \label{fig:tsne_text_image}
  \end{minipage}
  % \vspace{-6pt}
\end{figure}

To further illustrate our motivation to use text classifier for label transfer, we show the tSNE visualizations of the feature embeddings derived from a source-trained sentence classifier, and compare them to the features derived from a source-trained image classifier in \cref{fig:tsne_text_image}. Evidently, the features computed using the text classifier (\cref{fig:geoimnet-text,fig:geoplaces-text}) are well-separated (more intra-class separation) and well-aligned (less inter-domain separation) compared to image classifier (\cref{fig:geoimnet-image,fig:geoplaces-image}) further validating our hypothesis that the text descriptions of same-class images from both within and across domains lie close to each other.
% While the source labels provide strong supervision during training the images, the target psuedo-labels derived from text-modality offer noisy yet highly valuable training signal pertaining to the target domain, helping improve the overall performance on target images post transfer. 

%%%%%%%%%%%%%%%%%%%%%%%%%%%%%%%%%%%%%%%%%%%%%%%%%%%%%%%%%%%%%%%%%%%%%%%%%%%%%%%%%%%%%%%%%%%%%%%%%%%%%
%%%%%%%%%%%%%%%%%%%%%%%%%%%%%%%%%%%%%%%%%%%%%%%%%%%%%%%%%%%%%%%%%%%%%%%%%%%%%%%%%%%%%%%%%%%%%%%%%%%%%
%%%%%%%%%%%%%%%%%%%%%%%%%%%%%%%%%%%%%%%%%%%%%%%%%%%%%%%%%%%%%%%%%%%%%%%%%%%%%%%%%%%%%%%%%%%%%%%%%%%%%
%%%%%%%%%%%%%%%%%%%%%%%%%%%%%%%%%%%%%%%%%%%%%%%%%%%%%%%%%%%%%%%%%%%%%%%%%%%%%%%%%%%%%%%%%%%%%%%%%%%%%

% \vspace{-12pt}
\noindent \para{Cross-modal supervision transfer.} We distill the powerful discriminative knowledge learned from text into images through cross-modal (text to image) supervision transfer in the target domain. Specifically, we first freeze the weights of the source-trained BERT classifier $\mathcal{B}$ and compute pseudo-labels on all the target images using their corresponding text descriptions. For an image $x_i^t$ with caption $c_i^t$,
% %
\begin{equation}
    \hat{y}_i^t = \argmax_{\mathcal{C}} \mathcal{B}(c_i^t; \phi^*),
    \label{eq:pseudo}
\end{equation}
% %
where $\mathcal{C}$ is the set of categories in the classification task. Using these predictions, we construct a pseudo-labeled target dataset, given by
% These predictions are then used as pseudo-labels for the unlabeled target images, to obtain pseudo-label augmented target dataset 
$\widehat{\mathcal{D}_t} = \{x_i^t, \hat{y}_i^t\}_{i=1}^{N_t}$. Finally, we combine this pseudo-labeled target images along with manually labeled source domain images to train an image classifier backbone $\mathcal{G}$ by sampling an equal number of images from both source and target in each mini-batch to eliminate effects caused by different dataset sizes. 
\begin{IEEEeqnarray}{C}
    \argmin_\theta \mathop{\mathbb{E}}_{(x_i,y_i) \sim \mathcal{D}_s} \mathcal{L}_{\text{CE}} (\mathcal{G}(x_i;\theta), y_i)  + \nonumber \\
                            \mathop{\mathbb{E}}_{(x_i,\hat{y}_i) \sim \mathcal{\widehat{D}}_t} \mathcal{L}_{\text{CE}} (\mathcal{G}(x_i;\theta), y_i) 
    \label{eq:joint_training}
\end{IEEEeqnarray}
Note that the inference is performed exclusively using the trained image-based classifier $\mathcal{G}(;\theta^*)$ on image inputs, and neither the text inputs nor the sentence-classifier $\mathcal{B}$ is needed or used at test-time. 

\subsection{Extending \Ours{} to Handle Outliers}
\label{sec:source_free_pseudo}

Owing to the simplicity in the design, \Ours{} can easily be extended to the case where the target domain potentially contains outlier samples from outside the category set, also called open-world or universal adaptation (UniDA)~\cite{you2019universal, saito2020universal}.
% with minimum modifications to the framework. 
% or source-free unsupervised domain adaptation settings 
% with minimum modifications to the framework. 
While classical transfer necessitates complete matching between source and target category spaces, open-world transfer relaxes this requirement, allowing the possibility of encountering images from previously unseen and outlier categories during test-time in the target domain~\cite{you2019universal, saito2020universal}. The task is then to accurately classify a test-image into one of $\mathcal{C}_s$ categories shared between source and target domains while simultaenously detecting outlier images from target private classes. To suit \Ours{} for UniDA, we modify \cref{eq:pseudo} to additionally label predictions made by the text-classifier network $\mathcal{B}$ with an \textit{outlier} class using maximum softmax probability thresholding~\cite{hendrycks2016baseline} after training. 
\begin{equation}
    \hat{y}_i^t =
    \begin{cases}
        \argmax_{\mathcal{C}_s} \mathcal{B}(c_i^t; \phi^*) & \text{if } \max_{\mathcal{C}_s} \mathcal{B}(c_i^t; \phi^*) > \tau \\
        |\mathcal{C}_s| + 1 & \text{otherwise},
    \end{cases}
\end{equation}
where $\tau$ is a threshold used to detect outlier samples during inference. We then proceed to train a downstream classifier on $|\mathcal{C}|_s{+}1$ classes using data from supervised source and psuedo-labeled target data from shared classes as well as the outlier class. During inference, we assign a test-image to one of the $\mathcal{C}_s$ classes or the special {outlier} class based on the prediction. We heuristically choose $\tau=0.75$ and do not ablate on this. We show in \cref{subsec:unida} that this simple extension yields highest accuracy on the challenging GeoUniDA dataset~\cite{kalluri2023geonet}, highlighting the versatility of \Ours{} to handle diverse styles of domain transfer. 

\section{Experiments}
\label{sec:expmnts}

% We first present the effectiveness of \Ours{} on image classification (\cref{subsec:images}) and extensions to open world transfer (\cref{subsec:unida}) followed by results on a new dataset for transfer between ego-exo views in videos (\cref{subsec:videos}) and extensive ablations into our framework (\cref{subsec:ablation}). 

\subsection{\Ours{} for Image Classification}
\label{subsec:images}

\para{Datasets.} We adopt GeoNet~\cite{kalluri2023geonet} and DomainNet~\cite{peng2019moment} datasets which together cover a range of domain shifts across varying difficulty levels.
% validating our effectiveness across a diverse set of challenging scenarios.
GeoNet is the largest dataset for domain adaptation with more than 750k images, proposed to study a practical real-world problem of geographic disparities in images for two tasks - GeoImnet for image classification from 600 classes and GeoPlaces for scene recognition from 205 classes. 
% GeoNet also provides a benchmark called GeoUniDA for evaluating universal adaptation capability, 
% which are the most complicate
% d and challenging settings studying adaptation. 
DomainNet is a challenging dataset proposed for adaptation with 400,000 images from 345 classes. Following prior work~\cite{wei2021toalign, kalluri2022memsac}, we show our results on all 12 transfer settings from the 4 most studied domains  \textit{real, clipart, sketch} and \textit{painting}.
% We use the publicly available train-test splits from both these datasets in our training and evaluation. 
We use a ViT-base~\cite{dosovitskiy2020image} backbone as the image encoder on the GeoNet, and follow prior work~\cite{zhu2023patch} and use Swin-base backbone~\cite{liu2021swin} for experiments on DomainNet. Complete training details are included in the \supp{} (\cref{sec:training_details}).

% These benchmarks together cover a suite of domain shifts across varying difficulty levels, which helps validate our approach across a diversity of challenhging scenarios. 
% GeoYFCC is another dataset proposed for geographic adaptation, with 75k images from 68 categories across two domains. 

% \vspace{-6pt}
\para{Source of text supervision.} For GeoNet, we use text supervision from the metadata publicly released along with the dataset, and concatenate the tags, alt-text and free-form captions provided for each image to create the text descriptions. For the DomainNet dataset, since no associated text descriptions are provided, we use a BLIP-2~\cite{li2023blip2} model to generate short captions for each image from all the domains. Note that our method only requires text during training, and inference is done solely based on images. 
% Sample text descriptions for images from both datasets are shown in the \supp{}. 
% Note that these metadata or captions can be potentially noisy and unstructured with high variance in quality and coherence, posing an additional challenge in leveraging these inputs for discriminative transfer. 

% \para{Training Details} 
% \vspace{-6pt}
\para{Baselines.} A possible argument for the effectiveness of text supervision might be the direct presence of label information in the text description, eliminating the need for any manual supervision at all. To study this in greater detail, we devise two strong baselines to derive psuedo-labels directly using the text descriptions in the target without using any source domain data as follows. We first use a pre-trained Sentence-BERT~\cite{reimers-2019-sentence-bert} encoder, and compute the {label embeddings} of all the category names as $\mathbf{L} \in \mathbb{R}^{|\mathcal{C}| \times d}$, where $d$ is the embedding dimension of the sentence encoder, followed by zero-shot inference using:
(i) {\bf TextMatch,} where we compute the embedding of each text description $e_i^t \in \mathbb{R}^{1 \times d}$ from the target domain, and assign psuedo-label to the label with the highest similarity score with the text embeddings: $\hat{y}_i = \argmax_{|\mathcal{C}|} (e_i^t \cdot \mathbf{L}^T)$, and (ii) {\bf nGramMatch,} where we additionally compute the set of all $n$-grams $\{w\}$ for each text description $c_i$ for $n=\{1,2,3,4\}$ and find the embeddings for each of these ngrams separately: $\mathbf{W} \in \mathbb{R}^{|w| \times d}$. The pseudo-label is then assigned to the label with the highest similarity score with the best matching ngram: $\hat{y}_i = \argmax_{|\mathcal{C}|} \max_{w} (\mathbf{W} \cdot \mathbf{L}^T)$. Once the psuedo-labels are generated, we proceed with training a joint model using \cref{eq:joint_training} as before. 

In addition to these, we also compare the zero-shot classification obtained using \textbf{CLIP}~\cite{radford2021learning} with ViT-base backbone. We adopt domain-aware prompting following prior work~\cite{dunlap2023using, liu2023tdg}, where we incorporate the domain information into the prompt-text (eg: \texttt{A \textit{sketch} of a <class>} instead of \texttt{A \textit{photo} of a <class>} to classify sketch images).

% \para{Baselines and Prior Methods} We compare \Ours{} with several state-of-the-art methods from UDA literature,

\paragraph{\Ours{} Outperforms UDA on GeoNet}
% \label{subsec:geonet_results}

\begin{table}[!tbp]
  \centering
  \resizebox{.5\textwidth}{!}{%
  % \begin{tabular}{@{}lccccccccccccc@{}}
  \begin{tabular}{>{\columncolor{white}[0pt][\tabcolsep]}lccccccccccccc}
    \specialrule{2\heavyrulewidth}{0pt}{2pt}
    & \multicolumn{2}{c}{GeoImnet} && \multicolumn{2}{c}{GeoPlaces} && \multirow{2}{*}{Average} \\
    \cline{2-3} \cline{5-6} 
    & U$\rightarrow$A & A$\rightarrow$U && U$\rightarrow$A & A$\rightarrow$U \\
    \cmidrule(lr){1-8}
    \multicolumn{4}{l}{\textit{Unsupervised Adaptation}} \\
    Source Only & 52.46 & 51.91 && 44.90 & 36.85 && 46.53 \\
    % DANN & 50.68 & 51.73 && 40.80 & 32.99 && 44.05 \\
    CDAN~\cite{CDAN} & 54.48 & 53.87 && 42.88 & 36.21 && 46.86 \\
    MemSAC~\cite{kalluri2022memsac} & 53.02 & 54.37 && 42.05 & 38.33 && 46.94 \\
    ToAlign~\cite{wei2021toalign} & 55.67 & 55.92 && 42.32 & 38.40 && 48.08 \\
    MDD~\cite{zhang2019bridging} & 51.57 & 50.73 && 42.54 & 39.23 && 46.02 \\
    DALN~\cite{Chen_2022_CVPR} & 55.36 & 55.77 && 41.06 & 40.41 && 48.15 \\
    PMTrans~\cite{zhu2023patch} & \underline{56.76} & \underline{57.60} && 46.18 & 40.33 && 50.22 \\
    \cmidrule(lr){1-8}
    \multicolumn{4}{l}{\textit{Zeroshot Classification}} \\
    % CLIP~\cite{radford2021learning} & 47.37 & 51.38 && 46.97 & 53.01 && 49.68 \\
    CLIP$^\dagger$~\cite{radford2021learning} & 49.84 & 53.83 && 43.41 & \underline{54.34} && 50.36 \\
    \cmidrule(lr){1-8}
    % \multicolumn{4}{l}{\textit{Zero-shot Text-based Labeling}} \\
    TextMatch  & 49.68 & 54.82 && \underline{53.06} & 50.11 && \underline{51.92} \\
    nGramMatch & 49.53 & 51.02 && 51.70 & 49.87 && 50.93 \\
    \cmidrule(lr){1-8}
    \rowcolor{gray!20} % Color for a specific row
    \Ours{} & \textbf{63.67} & \textbf{64.16} && \textbf{56.14} & \textbf{57.02} && \textbf{60.24} \\
    % \midrule
    % \textcolor{gray}{Target Supervised} & \textcolor{gray}{68.41} & \textcolor{gray}{68.15} && \textcolor{gray}{58.99} & \textcolor{gray}{63.57} && \textcolor{gray}{64.78} \\
    \specialrule{\heavyrulewidth}{2pt}{2pt}
  \end{tabular}%
  }
  \caption{\Ours{} outperforms all prior methods by ${>}10\%$ on average with the challenging GeoImnet benchmark with 600 classes and GeoPlaces with 205 classes designed for geographical transfer. All methods use a ViT-B backbone. $^\dagger$denotes domain aware-prompting. Best values in \textbf{bold}, second best \underline{underlined}. U:USA, A:Asia.}
  % \caption{Comparison of different methods using ViT-B/16 architectures. $\dagger$Domain-aware prompting.}
  \label{tab:geoDa}
  % \vspace{-6pt}
\end{table}

% %%%%%%%%%%%%%%%%%%%%%%%%%%%%%%%%%%%%%%%%%%%%%%%%%%%%%%%%%%%%%%%%
% %%%%%%%%%%%%%%%%%%%%%%%%%%%%%%%%%%%%%%%%%%%%%%%%%%%%%%%%%%%%%%%%
% %%%%%%%%%%%%%%%%  DomainNet-345 %%%%%%%%%%%%%%%%%%%%%%%%%%%%%%%%
% %%%%%%%%%%%%%%%%%%%%%%%%%%%%%%%%%%%%%%%%%%%%%%%%%%%%%%%%%%%%%%%%

\begin{table*}[!tbp]
    \centering
    \setlength{\tabcolsep}{3pt}
    % \rowcolors{3}{gray!10}{white}
    \resizebox{.85\textwidth}{!}{
    % \begin{tabular}{@{} l *{17}{c} @{} }
    \begin{tabular}{>{\columncolor{white}[0pt][\tabcolsep]}lcccccccccccccccccccc}
        \specialrule{2\heavyrulewidth}{0pt}{2pt}
        Source & \multicolumn{3}{c}{\textbf{Real$\rightarrow$}} && \multicolumn{3}{c}{\textbf{Clipart$\rightarrow$}} && \multicolumn{3}{c}{\textbf{Sketch$\rightarrow$}} && \multicolumn{3}{c}{\textbf{Painting}$\rightarrow$} && \\
        \cline{2-4} \cline{6-8} \cline{10-12} \cline{14-16}
        Target & {C} & {S} & {P} && {R} & {S} & {P} && {R} & {C} & {P} && {R} & {C} & {S} && Avg. \\
        \cmidrule(lr){1-18}
        \multicolumn{7}{l}{\textit{Unsupervised Adaptation}} \\
        Source Only & 63.02 & 49.47 & 60.48 && 70.52 & 56.09 & 52.53 && 70.42 & 65.91 & 54.47 && 73.34 & 60.09 & 48.25 && 60.38 \\
        MCD~\yrcite{saito2018maximum} & 39.40 & 25.20 & 41.20 && 44.60 & 31.20 & 25.50 && 34.50 & 37.30 & 27.20 && 48.10 & 31.10 & 22.80 && 34.01 \\
        MDD~\citep{zhang2019bridging} & 52.80 & 41.20 & 47.80 && 52.50 & 42.10 & 40.70 && 54.20 & 54.30 & 43.10 && 51.20 & 43.70 & 41.70 && 47.11 \\
        CGDM~\cite{du2021cross} & 49.40 & 38.20 & 47.20 && 53.50 & 36.90 & 35.30 && 55.60 & 50.10 & 43.70 && 59.40 & 37.70 & 33.50 && 45.04 \\
        SCDA~\cite{li2021semantic} & 54.00 & 42.50 & 51.90 && 55.00 & 44.10 & 39.30 && 53.20 & 55.60 & 44.70 && 56.20 & 44.10 & 42.00 && 48.55 \\
        SSRT-B~\cite{sun2022safe} & 69.90 & 58.90 & 66.00 && 75.80 & 59.80 & 60.20 && 73.20 & 70.60 & 62.20 && 71.40 & 61.70 & 55.20 && 65.41 \\    
        MemSAC~\cite{kalluri2022memsac} & 63.49 & 42.14 & 60.32 && 72.33 & 54.92 & 46.14 && 73.46 & 68.04 & 52.75 && 74.42 & 57.79 & 43.57 && 59.11 \\
        CDTrans~\cite{xu2021cdtrans} & 66.20 & 52.90 & 61.50 && 72.60 & 58.10 & 57.20 && 72.50 & 69.00 & 59.00 && 72.10 & 62.90 & 53.90 && 63.16 \\
        PMTrans~\cite{zhu2023patch} & \underline{74.10} & 61.10 & \textbf{70.00} && 79.30 & 63.70 & 62.70 && 77.50 & \underline{73.80} & 62.60 && 79.80 & 69.70 & 61.20 && 69.63 \\
        \cmidrule(lr){1-18}
        \multicolumn{7}{l}{\textit{Zero-shot Classification}} \\
        % CLIP~\cite{radford2021learning} & 64.93 & 57.27 & 61.88 && 81.26 & 57.27 & 61.88 && \underline{81.26} & 64.93 & 61.88 && \underline{81.26} & 64.93 & 57.27 && 66.34 \\
        CLIP$^\dagger$~\cite{radford2021learning} & 72.39 & 60.90 & 66.81 && \textbf{81.37} & 60.90 & \underline{66.81} && \textbf{81.37} & 72.39 & \underline{66.81} && \textbf{81.37} & \underline{72.39} & 60.90 && 70.38 \\
        \cmidrule(lr){1-18}
        % \multicolumn{7}{l}{\textit{Zero-shot Text-based Labeling}} \\
        TextMatch & 71.36 & \underline{64.30} & 65.32 && 81.25 & \underline{65.65} & 64.85 && \underline{81.09} & 72.65 & 63.94 && \underline{81.08} & 70.84 & \textbf{64.17} && 70.14  \\
        nGramMatch & 68.92 & 59.82 & 63.15 && 76.35 & 61.72 & 62.87 && 76.35 & 69.28 & 62.51 && 76.04 & 68.52 & 60.52 && 67.17 \\
        \cmidrule(lr){1-18}
        \rowcolor{gray!20} % Color for a specific row
        \Ours{} & \textbf{77.30} & \textbf{68.25} & \underline{67.35} && \underline{81.31} & \textbf{67.03} & \textbf{66.81} && 80.78 & \textbf{75.62} & \textbf{68.08} && 79.23 & \textbf{73.80} & \underline{63.44} && \textbf{72.41} \\
        % \midrule
        % \textcolor{lightgray}{Tgt. Supervised} & \textcolor{lightgray}{72.59} & \textcolor{lightgray}{62.66} &\textcolor{lightgray}{65.12} & & \textcolor{lightgray}{80.92} & \textcolor{lightgray}{62.66} & \textcolor{lightgray}{65.12} & & \textcolor{lightgray}{80.92} & \textcolor{lightgray}{72.59} & \textcolor{lightgray}{65.12} & & \textcolor{lightgray}{80.92} & \textcolor{lightgray}{72.59} & \textcolor{lightgray}{62.66 } && \textcolor{lightgray}{70.32} \\
        \specialrule{\heavyrulewidth}{2pt}{2pt}
    \end{tabular}
    }
    \captionsetup{width=\textwidth}
    \caption{\Ours{} sets new state-of-the-art on DomainNet-345 dataset, outperforming prior methods and baselines in most tasks. All models use Swin-B backbone, and UDA numbers are taken from \cite{zhu2023patch}. $^\dagger$denotes domain aware-prompting. Best values in \textbf{bold}, second best \underline{underlined}. \\
    R:Real, C:Clipart, S:Sketch, {P:Painting}.}
    \label{tab:domainnet345}
    % \vspace{-8pt}
\end{table*}

% We present results for GeoPlaces and GeoImnet benchmarks in \cref{tab:geoDa}. 
As noted in \cite{kalluri2023geonet}, previous UDA methods often fall short of bridging geographic disparities, highlighting the challenge of geographical transfer with image data alone. From \cref{tab:geoDa}, \Ours{} achieves $60.24\%$ average Top-1 average accuracy on GeoNet, beating all previous UDA methods and strong baselines by significant margins, providing solid validation to our transfer approach using language guidance. Specifically, \Ours{} outperforms the source-only baseline by ${\sim}14\%$ and best adaptation approach PMTrans~\cite{zhu2023patch} by ${\sim}10\%$ on the average accuracy, highlighting the natural benefit conferred by training while leveraging text supervision in source and target domains. \Ours{} even surpasses zeroshot accuracy using domain-aware prompting on CLIP~\cite{radford2021learning} by ${\sim}10\%$, while being trained on order of magnitude fewer data compared to CLIP's hundreds of millions of image-text pairs. 
Remarkably, we also outperform the strongest baseline \textit{TextMatch} by ${\sim}8\%$, underlining the fact that in cases when the text descriptions might not always have embedded label information directly, using labels from a source is still advantageous. 

\paragraph{\Ours{} achieves new SOTA on DomainNet} 
We summarize the results on DomainNet in \cref{tab:domainnet345}, where \Ours{} yields large gains over several prior UDA methods and all the competitive baselines, setting new state-of-the-art. Notably, many prior methods return lesser numbers than directly training on a source model~\cite{saito2018maximum, zhang2019bridging, du2021cross, li2021semantic}, indicating their poor scalability to natural domain shifts in large-scale data. While more recent innovations in UDA such as self-training~\cite{sun2022safe} and patch-based mixing~\cite{zhu2023patch} offer improved performance, \Ours{} still outperforms these methods on most tasks. Finally, our superior accuracy compared to both baselines \textit{TextMatch} and \textit{nGramMatch}, that employ target-only pseudo-labeling, underscores the significance of having access to supervised text data and labels from a source domain for enhanced target accuracy. Further, a potential explanation for limited benefits being observed in LaGTran, as well as all previous UDA methods, when compared to CLIP under non-real to real transfer could be the precise match between images from standard categories in the real-target domain and the multi-million-scale training data utilized in CLIP. This alignment potentially eliminates any significant domain shift between the train and test settings, unlike in \Ours{} where we train on non-real images yet achieve accuracies that are competitive to CLIP. Notably, LaGTran still outperforms CLIP on all domains in GeoNet (+10\%) and most domains in DomainNet (upto +7\%) while being trained on multiple-orders of magnitude lesser image-text pairs than CLIP.

%%%%%%%%%%%%%%%%%%%%%%%%%%%
%%%%%%%%%%%%%%%%%%%%%%%%%%%
%%%%% GeoUniDA.   %%%%%%%%%
%%%%%%%%%%%%%%%%%%%%%%%%%%%

\begin{table}[!tbp]
\centering
\setlength{\extrarowheight}{2pt} % adjust the height of the row
\resizebox{0.47\textwidth}{!}{%
\begin{tabular}{>{\columncolor{white}[0pt][\tabcolsep]}l c c c} % Three columns, the first one is left-aligned and the next two are centered
    \toprule
     Method & Closed Set Acc. & Open Set Acc. & H-score \\
    \cmidrule(lr){1-4}
    Source Only w/MSP                & 38.00 & 73.90 & 50.20 \\
    UniDA~\cite{you2019universal}    & 27.64 & 43.93 & 33.93 \\
    DANCE~\cite{saito2020universal}  & 38.54 & 78.73 & \underline{51.75} \\
    OVANet~\cite{saito2021ovanet}    & 36.54 & 66.89 & 47.26 \\
    \cmidrule(lr){1-4}
    \rowcolor{gray!20} % Color for a specific row
    \Ours{}                   & 52.98 & 72.35 & \textbf{61.16} \\
    \bottomrule
\end{tabular}
}
\caption{{\bf Results on open-world transfer on GeoUniDA} shows strong performance of \Ours{} even with target outlier classes, achieving the highest H-score. Baseline numbers takes from \cite{kalluri2023geonet}. }
\label{tab:geounida}
% \vspace{-12pt}
\end{table}
\subsection{\Ours{} Improves Transfer with Outliers}
\label{subsec:unida}

We show our results on open-world transfer setting using the GeoUniDA dataset~\cite{kalluri2023geonet}, which examines unsupervised transfer across geographies in the presence of geographically unique classes in both source and target along with common classes. Specifically, GeoUniDA contains 62 shared classes between source and target, along with 138 private categories in each domain. 
% We train our universal adaptation method using the training procedure outlined in \cref{sec:source_free_pseudo}, 
We follow OVANet~\cite{saito2021ovanet} to adopt the H-score evaluation metric, which gives equal importance to closed-set and open-set accuracies by measuring the harmonic mean of both. In addition to standard works that address outlier detection through universal adaptation~\cite{you2019universal, saito2021ovanet, saito2020universal}, we also train a baseline model using only the source domain data performing test-time outlier detection using MSP thresholding~\cite{hendrycks2016baseline}. As shown in \cref{tab:geounida}, \Ours{} achieves a H-score of $61.16\%$, significantly surpassing the baseline source-only accuracy as well as all prior universal adaptation approaches by ${>}10\%$, indicating that language guidance naturally provides a strong signal to detect target samples while handling outliers in open-set target domain data. 

\subsection{\Ours{} for Video Domain Adaptation}
\label{subsec:videos}

% To study the benefit of language in improving cross-domain transfer beyond images, we create and study a new video adaptation dataset for transfer between ego-centric (first-person) and exo-centric (third-person) videos. We next introduce the dataset and show the strong performance of \Ours{} on this challenging and novel transfer setting.

\para{\egoexo{} dataset.} 
% The current benchmarks been proposed in prior literature for video domain adaptation~\cite{chen2019temporal, munro2020multi, plizzari2023cook} 
%
% Current benchmarks for domain adaptation in video are either saturated in accuracy (${>}95\%$ accuracy on HMDB$\leftrightarrow$UCF)~\cite{chen2019temporal}, limited in scale~\cite{choi2020unsupervised} or restrictive in domain shifts studied (different kitchens in \cite{munro2020multi}). As a result, 
Despite rapid advances in methods~\cite{chen2019temporal, munro2020multi, choi2020unsupervised, wei2022unsupervised} and benchmarks~\cite{munro2020multi, plizzari2023cook} for video domain adaptation, little insight is available into their ability to address challenging settings such as transfer between ego (first-person) and exo (third-person) perspectives in videos. While prior efforts studying ego-exo transfer require paired videos from both views~\cite{quattrocchi2023synchronization, sigurdsson2018actor, huang2024egoexolearn} or do not leverage target unlabeled data~\cite{li2021ego, ohkawa2023exo2egodvc}, limited works study unsupervised domain transfer from ego to exo views due to the lack of a suitable benchmark. 

\begin{figure}[!tpb]
  \begin{minipage}[b]{0.23\textwidth}
    \centering
    \includegraphics[width=\linewidth]{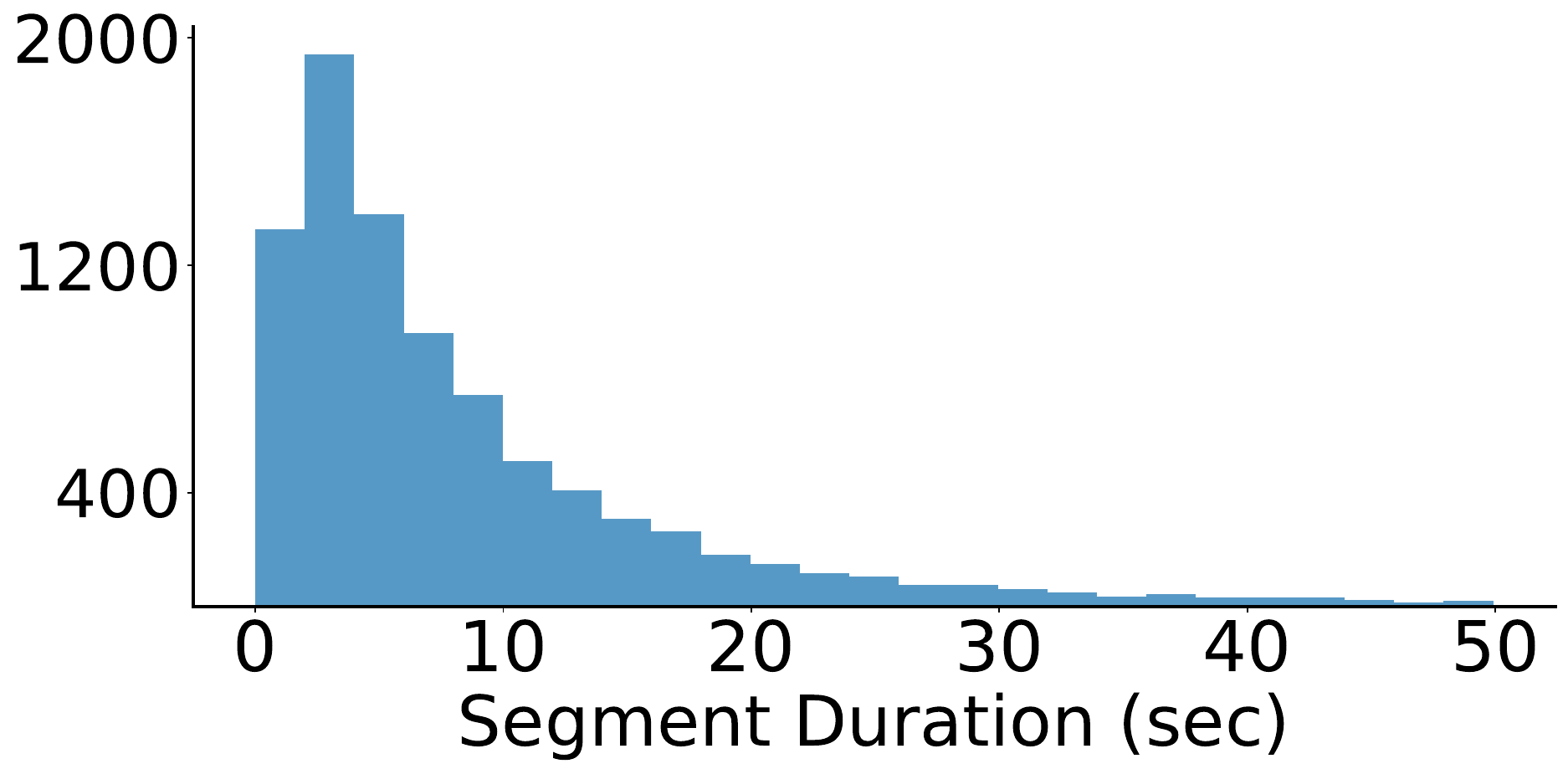}
    \subcaption{Segment Duration}
    \label{fig:segment_duration}
  \end{minipage}
  ~
  \begin{minipage}[b]{0.23\textwidth}
    \centering
    \includegraphics[width=\linewidth]{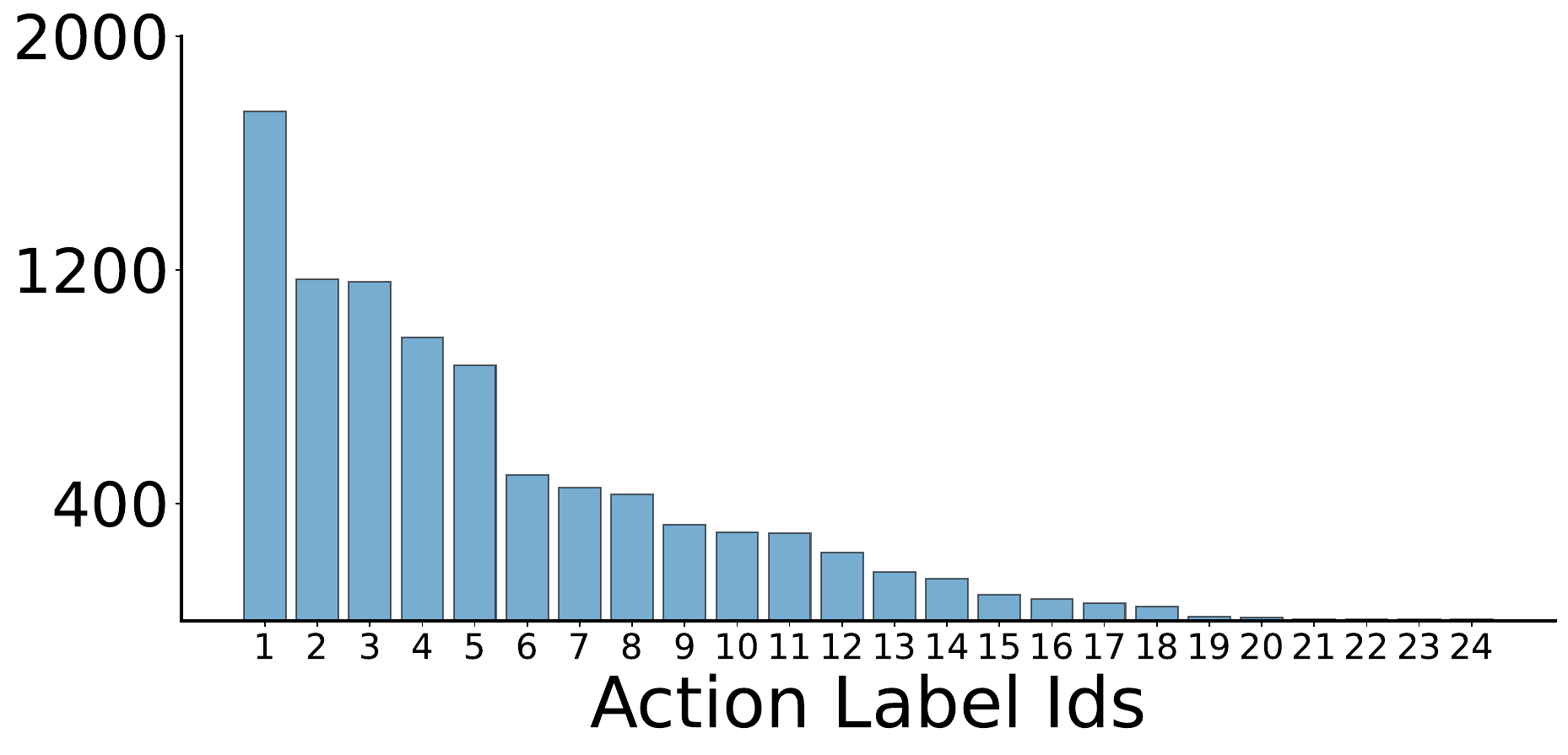}
    \subcaption{Label Distribution}
    \label{fig:label_distribution}
  \end{minipage}
  \caption{{\bf Dataset Statistics for \egoexo{}:} \cref{fig:segment_duration} Shows the distribution of segment durations of action videos from \egoexo{} which range from 0.4sec-1min. \cref{fig:label_distribution} shows the long-tail of category distribution in \egoexo{} indicating the challenge in robust classification and transfer. }
  \label{fig:egoexo_stats}
  % \vspace{-6pt}
\end{figure}

Therefore, we introduce a new benchmark called \egoexo{} to study transfer between the ego and exo views in videos. 
We curate our dataset using the recently proposed Ego-Exo4D~\cite{grauman2023ego}, utilizing their keystep annotations for action labels, and atomic descriptions as text supervision. We manually remap the labels to a coarser hierarchy to ease the difficult task of predicting very fine-grained action classes from short clips (eg: \texttt{add coffee beans} vs. \texttt{add coffee grounds}). Complete details about our dataset creation process are included in \cref{sec:egoexo_creation}. 
Our proposed \egoexo{} consists of video segments labeled with actions from one of the 24 keysteps from ego and exo views of the corresponding actions. We obtain 4100 ego-videos and 4986 exo-videos capturing variety of actions and scenes. The atomic action descriptions from all the timestamps within each segment form the text supervision for that segment. 
The same procedure is applied to the validation videos yielding 3147 segments with both ego and exo views. The distribution of the duration of segments in the benchmark, along with the label distribution for ego and exo domains is presented in \cref{fig:egoexo_stats}. 
We provide more details about the construction of the dataset in the \supp{} material. 
The videos, labels along with the text descriptions are publicly available on our project page.
% \url{https://tarun005.github.io/lagtran/}.
% will be publicly released to foster future research, and are also included with the \supp{}. 

%%%%%%%%%%%%%%%%%%%%%%%%%%%%%%%%%%%%%%%%%%%%%%%%%%%%%
%%%%%%%%%%%%%%%%%%%%%%%%%%%%%%%%%%%%%%%%%%%%%%%%%%%%%
%%%%%%%%%%%%%%%%%%%%%%%%%%%%%%%%%%%%%%%%%%%%%%%%%%%%%
%%%%%%%%%%%%%%%%%%%%%%%%%%%%%%%%%%%%%%%%%%%%%%%%%%%%%

% \vspace{-8pt}
\para{Training details.} We use the pre-computed Omnivore-base~\cite{girdhar2022omnivore} features provided along with the EgoExo4D dataset for training and evaluation, and follow the same strategy for training all the other baselines as well as prior adaptation methods for fair comparison. We use the top-1 accuracy on the validation set for evaluation. More details on the training procedure are provided in the \supp{}, \cref{sec:training_details}. We compare \Ours{} for video with prior UDA approaches~\cite{chen2019progressive, wei2022unsupervised} as well as Video-CLIP based methods with domain-aware prompting~\cite{lin2022egocentric, zhao2023learning}.
 
%%%%%%%%%%%%%%%%%%%%%%%%%%%
%%%%%%%%%%%%%%%%%%%%%%%%%%%
%%%%% Different ablation %%
%%%%%%%%%%%%%%%%%%%%%%%%%%%

\begin{table}[!tbp]
\centering
    \resizebox{0.45\textwidth}{!}{
        % \begin{tabular}{@{} l *{4}{c} @{} }
        \begin{tabular}{>{\columncolor{white}[0pt][\tabcolsep]}l c c c c} 
            \specialrule{2\heavyrulewidth}{0pt}{2pt}
            & Ego$\rightarrow$Exo & Exo$\rightarrow$Ego & Avg. \\
            \cmidrule(lr){1-4}
            \multicolumn{3}{l}{\textit{Unsupervised Adaptation}} \\
            Source Only & {8.39} & {15.66} & {12.03}  \\
            % CDAN~\cite{CDAN} & 14.08 & 17.03 & 15.56 \\
            % UAR~\cite{choi2020unsupervised} & 14.65 & 18.75 & 16.70 \\
            TA3N~\cite{chen2019temporal} & 6.92 & \underline{27.95} & 17.44 \\
            TransVAE~\cite{wei2022unsupervised} & \underline{12.06} & {23.34} & \underline{17.70} \\
            \cmidrule(lr){1-4}
            \multicolumn{3}{l}{\textit{Zero-shot Video Recognition}} \\
            EgoVLP~\cite{lin2022egocentric} & 5.89 & 19.35 & 12.62 \\
            LaVILA~\cite{zhao2023learning} & 5.86 & 23.16 & 14.51 \\
            \cmidrule(lr){1-4}
            % \multicolumn{3}{l}{\textit{Zero-shot Text-based Labeling}} \\
            TextMatch & 10.36 & 13.57 & 11.97 \\
            nGramMatch & 11.50 & 15.46 & 13.98 \\
            \cmidrule(lr){1-4}
            \rowcolor{gray!20} % Color for a specific row
            \Ours{} & \textbf{12.34} & \textbf{30.76} & \textbf{21.55} \\
            \cmidrule(lr){1-4}
            \textcolor{gray}{Target Sup.} & \textcolor{gray}{17.91} & \textcolor{gray}{33.19} & \textcolor{gray}{25.55} \\
            % \bottomrule
            \specialrule{\heavyrulewidth}{0pt}{2pt}
        \end{tabular}
        }
    \captionsetup{width=0.5\textwidth}
    \caption{{\bf Results on \egoexo{} benchmark} \Ours{} achieves the highest accuracy compared to prior video UDA methods as well as zeroshot video-text pre-trained models. Best values in \textbf{bold}, second best \underline{underlined}. All methods use pre-extracted omnivore-base features, EgoVLP and LaVILA use Timesformer-base backbone.} 
    \label{tab:egoexoda}
% \vspace{-8pt}

\end{table}
\para{\Ours{} efficiently handles cross-view transfer.} Firstly, we highlight the importance of studying robustness across ego and exo views in \cref{tab:egoexoda} by examining the ego-test accuracy of a model trained directly on ego videos, which achieves 33.19\%, compared to a model transferred from exo-videos, which only achieves 15.66\%. Similarly, a model trained on ego videos achieve only 8.4\% for recognition in exo view, compared to a potential 17.91\% achievable by training directly on exo videos, indicating a significant domain shift. Current state-of-the-art video adaptation methods~\cite{wei2022unsupervised} yield limited gains to bridge these gaps, highlighting the need for novel approaches to address this challenge. Moreover, zeroshot video classification accuracy using EgoVLP~\cite{lin2022egocentric} and LaVILA~\cite{zhao2023learning} also show limited gains. 
% , although trained on thousands of video-text pairs from Ego4d~\cite{grauman2022ego4d} dataset using a Timesformer-base backbone, similar in size to Omnivore-base. 
%
Notably, \Ours{} which efficiently leverages action descriptions available alongside the videos, achieves an accuracy of 21.55\% on average significantly outperforming the source-only baseline by $9\%$ and prior adaptation methods by ${>}4\%$. \Ours{} also outperforms pseudo-labeling using \textit{nGramMatch} or \textit{TextMatch}, as the text descriptions, independently developed from keystep labels, often lack utility for deciphering the action category labels on their own. We also note the substantial scope for further improvement in future, both in terms of the low within-domain accuracy as well as the remaining gap to supervised target accuracy.

\subsection{Analysis and Ablations}
\label{subsec:ablation}

% \vspace{-12pt}
\para{How much text supervision is needed for \Ours{}?}

\begin{figure}[!tbp]
  \begin{minipage}[b]{0.46\linewidth}
    \centering
    \includegraphics[width=\linewidth]{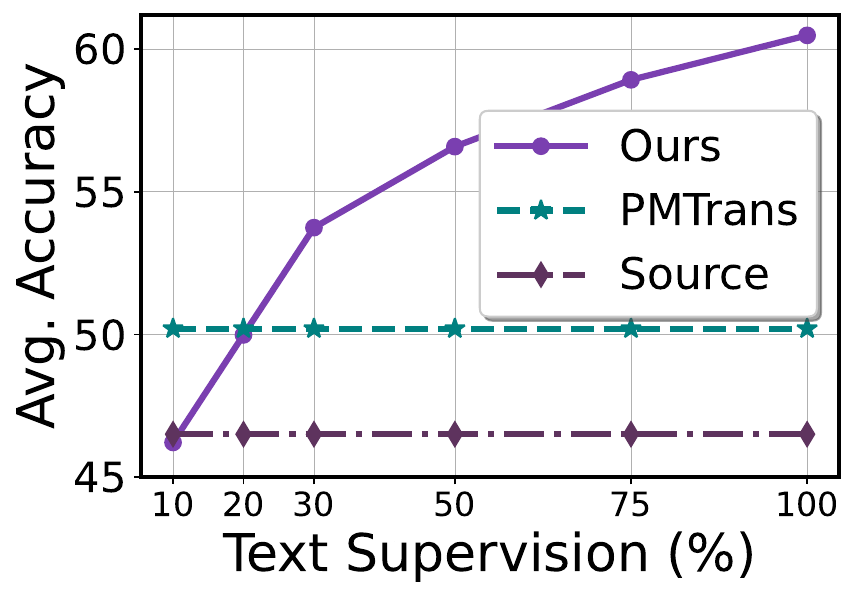}
    \subcaption{Accuracy on GeoNet}
    \label{fig:datavol_geonet}
  \end{minipage}
  ~~~
  \begin{minipage}[b]{0.46\linewidth}
    \centering
    \includegraphics[width=\linewidth]{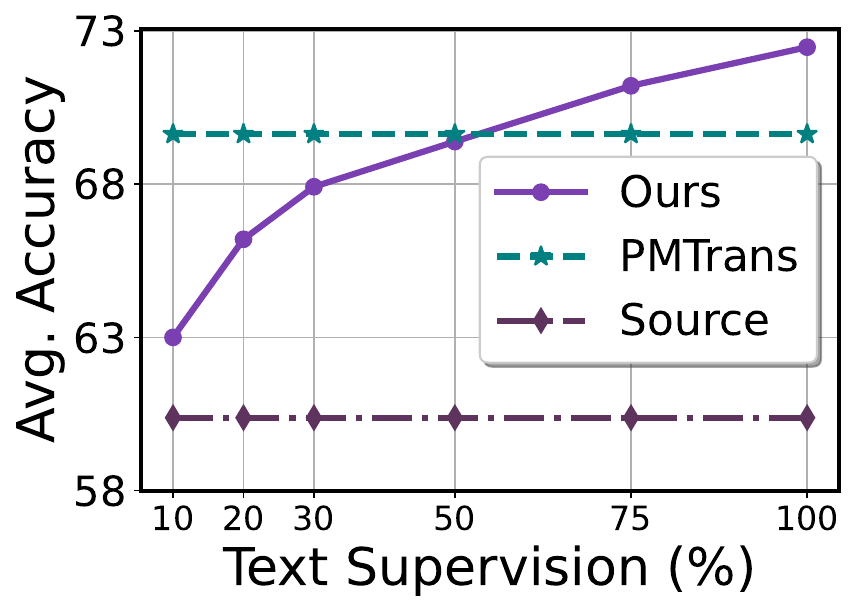}
    \subcaption{Accuracy on DomainNet}
    \label{fig:datavol_domainnet}
  \end{minipage}
  \caption{{\bf Impact of the amount of text supervision on the target accuracy.} \Ours{} outperforms strong UDA methods while requiring text supervision from only 20\% of samples in GeoNet and 50\% in DomainNet, with potential for further enhancement with increased text data.}
  \label{fig:datavol}
  % \vspace{-12pt}
\end{figure}

Since natural language supervision is fundamental to \Ours{}, we analyze the impact of the amount of supervision available on the eventual target accuracy. We retrain \Ours{} by assuming text supervision from only $\mu\%$ of images in both source and target domains, where $\mu=\{10,20,30,50,75,100\}\%$, and simply discard the target images that do not have corresponding textual supervision. As shown in \cref{fig:datavol}, \Ours{} outperforms image-only method PMTrans~\cite{zhu2023patch} even with just 20\% text supervision in GeoNet(~\cref{fig:datavol_geonet}) and 50\% in DomainNet(~\cref{fig:datavol_domainnet}), indicating its high data efficiency. Notably, the graph remains unsaturated, suggesting the potential for further improvement through the collection of more cheaply available text supervision in the target domain. 

% \vspace{-8pt}
\para{Effect of text classifier backbone.} 
We compare different choices of text classifiers such as DistilBERT~\cite{sanh2019distilbert}, T5-Small~\cite{raffel2020exploring}, GPT2~\cite{radford2019language} as well as text branch of CLIP~\cite{radford2021learning} (CLIP-T) using text-classification accuracy on our datasets.
We refer readers to the respective papers for details on their architectures and pre-training datasets. From \cref{tab:text_classifier}, DistilBERT yields best text-classification accuracy on all our three benchmarks, outperforming text-only models like T5 and GPT2. Despite large-scale vision-language pre-training, CLIP-T did not yield substantial benefits.

%%%%%%%%%%%%%%%%%%%%%%%%%%%
%%%%%%%%%%%%%%%%%%%%%%%%%%%
%%%%% Language Model %%%%%%
%%%%%%%%%%%%%%%%%%%%%%%%%%%

\begin{table}[!tbp]
\centering
    \resizebox{\linewidth}{!}{
        \begin{tabular}{@{} l *{8}{c}@{}}
            \toprule
            Model & params (M)  & GeoImnet & GeoPlaces & DomainNet \\
            \midrule
            T5-small~\cite{raffel2020exploring} & 60.87 & 73.93 & 63.61 & 68.57 \\
            CLIP-T~\cite{radford2021learning} & 63.16 & 79.87 & 66.45 & 71.15 \\
            GPT-2~\cite{radford2019language} & 124 & 77.88 & 66.65 & 69.60 \\
            DistilBERT~\cite{devlin2019bert} & 67.1 & \textbf{83.53} & \textbf{69.31} & \textbf{71.43} \\
            \bottomrule
        \end{tabular}
        }
    \captionsetup{width=0.48\textwidth}
    % \vspace{-4pt}
    \caption{{\bf Comparison of text-classifier backbones} using text-classification accuracy on GeoNet and DomainNet datasets. BERT backbone outperforms other text-pretrained backbones and vision-language pre-trained CLIP-T.} 
    \label{tab:text_classifier}
% \vspace{-8pt}

\end{table}

% \begin{figure*}[!tbp]
%   \begin{minipage}[b]{0.49\linewidth}
%     \centering
%     \includegraphics[width=\linewidth]{icml2024/figures/llm_example_pic_Part1.pdf}
%     % \subcaption{Accuracy on GeoNet}
%     % \label{fig:figure1}
%   \end{minipage}
%   \hfill
%   \begin{minipage}[b]{0.49\linewidth}
%     \centering
%     \includegraphics[width=\linewidth]{icml2024/figures/llm_example_pic_Part2.pdf}
%     % \subcaption{Accuracy on DomainNet}
%     % \label{fig:figure2}
%   \end{minipage}
%   \caption{}
% \end{figure*}

\begin{figure}[!t]
  \begin{minipage}[b]{0.95\linewidth}
    \centering
    \includegraphics[width=\linewidth]{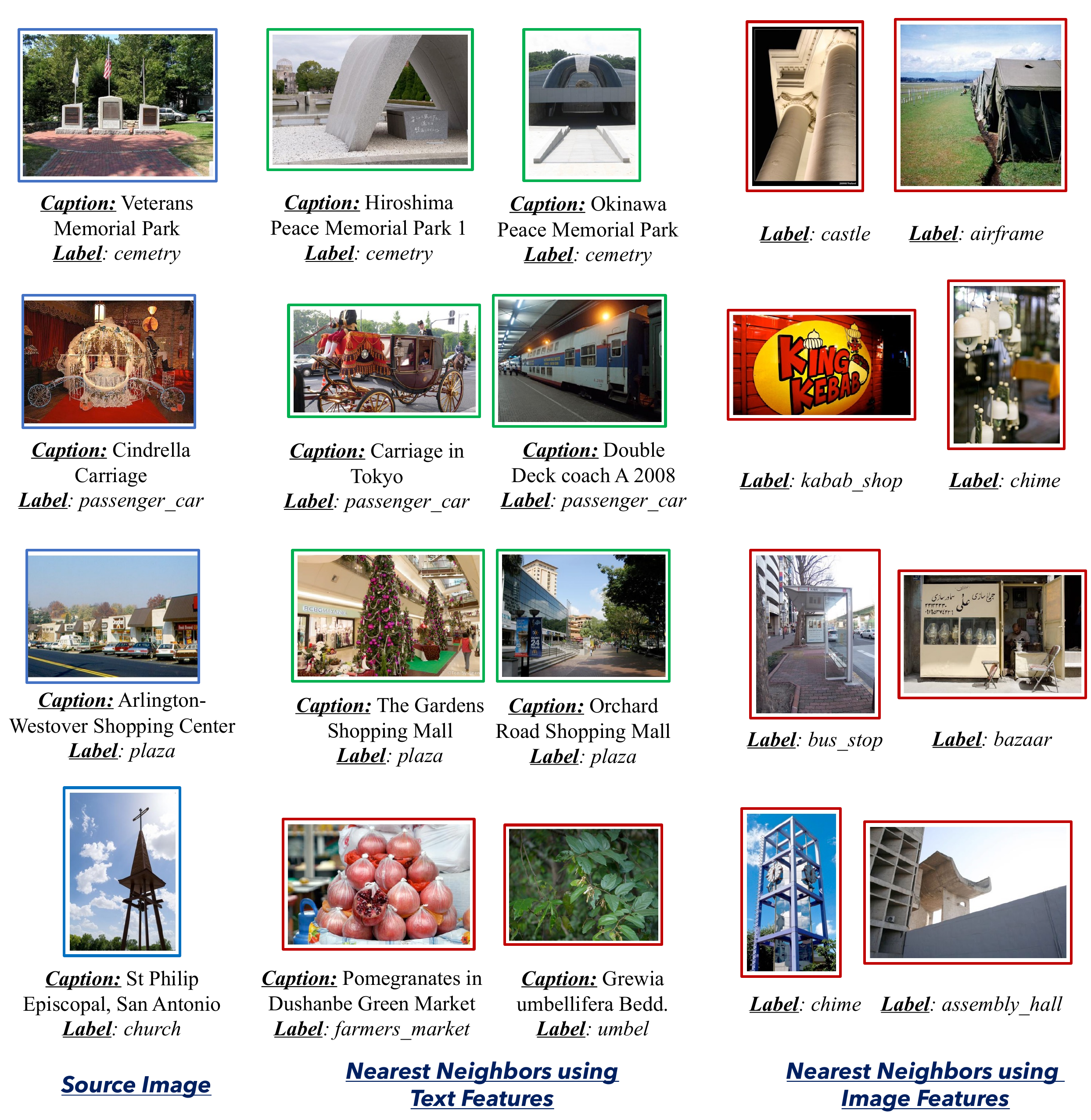}
    \caption{{\bf Visualization of nearest neighbors} of the leftmost source image, using text-trained and image-trained features, along with \textit{ground truth} labels for each image from GeoNet. We observe better ``same-class'' retrievals using text-captions due to reduced domain gap, as opposed to images. 
    % We also show a failure case of text in the last row, where both text and image features failed to retrieve images from the correct class ``church''.
    }
    \label{fig:knn_retrievals}
  \end{minipage}
  % \vspace{-15pt}
\end{figure}

% \vspace{-8pt}
\para{Importance of source domain images.} While the majority of our accuracy gains stem from the text guidance, the source domain images providing noise-free supervision are also important in learning strong models on the target domain. We observed that 
% From \cref{tab:source_data}, 
joint training using source and pseudo-labeled target yields improvements of $1.57\%$ for DomainNet and $0.8\%$ on \egoexo{} benchmarks compared to target-only training. More importantly, training jointly on source and target allows deploying a single joint model across both domains as opposed to domain specific models, greatly optimizing inference costs.

% text supervision is , we show in \cref{tab:} that source domain images also play a crucial role 

% \input{tables/source_data_ablation}

\vspace{-14pt}
\para{Nearest neighbors using image and text features.} We show the top-2 nearest neighbor retrievals using text-features computed from source-trained text-classifier as opposed to image-features in \cref{fig:knn_retrievals}. We observe more robust retrievals based on text-features corresponding to the captions of the images, rather than the images directly signifying the reduced domain gap in the text space. We also note a failure case in the last row of \cref{fig:knn_retrievals}, where neither the text features nor the image features could retrieve the image from the correct class \textit{church}. 
% \vspace{-8pt}

\section{Conclusion and Future Directions}

We introduce a novel framework called \Ours{} to use readily available text supervision and enhance target performance in unsupervised domain transfer scenarios. We first start with the observation that traditional domain alignment approaches yield limited benefits beyond well-understood domain shifts, followed by insights that language provides a semantically richer medium of transfer with reduced domain gaps. This leads to a language-guided transfer mechanism where we train a text classifier on language descriptions from a source domain and then use its predictions on descriptions from a different target domain as supervision for the corresponding images. Despite being conceptually simple and straightforward, we show the remarkable ability of our method to outperform competitive prior approaches on challenging benchmarks like GeoNet and DomainNet for images and proposed \egoexo{} for videos.
Through an emphasis on cost-effective or easily producible text supervision, we open new possibilities for advancing domain transfer in scenarios with limited manual supervision. 
Although \Ours{} achieves state-of-the-art performance across several datasets, it relies on external vision-language models for textual guidance in the absence of metadata, potentially constraining its applicability in scenarios where the pre-trained VLM models fail to offer discriminative text supervision. Additionally, while exhibiting fewer domain discrepancies, there remain non-trivial gaps even within the text modality that may reduce the accuracy of pseudo-labels in the target domain, which can be potentially addressed by additionally incorporating text-adaptation mechanisms.
% While \Ours{} already yields non-trivial boosts in performance, we identify potential areas of further improvement through additionally bridging the domain gaps in text space using language adaptation techniques~\cite{hung2023tada} or devising mechanisms to incorporate both image and language guidance that offer complimentary benefits during transfer.

\section*{Acknowledgements} 
We acknowledge support from NSF and a Google Award for Inclusion Research. 

\section*{Impact Statement}

Our paper presents an approach that can improve accuracy on domains facing label scarcity. Advancing this research area would 
enhance wider adoption of current AI technologies, and unlocks new capabilities in democratizing the progress in AI. 
Given that our proposed methodology only operates in the standard realm of image classification and our showcased results only use already publicly available datasets, we do not foresee any negative societal consequences specifically arising due to our method.

% This paper presents work whose goal is to advance the field of Machine Learning. There are many potential societal consequences of our work, none which we feel must be specifically highlighted here.

% In the unusual situation where you want a paper to appear in the
% references without citing it in the main text, use \nocite
% \nocite{langley00}

\bibliography{main}
\bibliographystyle{icml2024}

%%%%%%%%%%%%%%%%%%%%%%%%%%%%%%%%%%%%%%%%%%%%%%%%%%%%%%%%%%%%%%%%%%%%%%%%%%%%%%%
%%%%%%%%%%%%%%%%%%%%%%%%%%%%%%%%%%%%%%%%%%%%%%%%%%%%%%%%%%%%%%%%%%%%%%%%%%%%%%%
% APPENDIX
%%%%%%%%%%%%%%%%%%%%%%%%%%%%%%%%%%%%%%%%%%%%%%%%%%%%%%%%%%%%%%%%%%%%%%%%%%%%%%%
%%%%%%%%%%%%%%%%%%%%%%%%%%%%%%%%%%%%%%%%%%%%%%%%%%%%%%%%%%%%%%%%%%%%%%%%%%%%%%%
% \newpage
% \appendix
% \onecolumn

% swin encoder
% ego-exoDA dataset creation
% metadata available at test-time - pros and cons.
% source-free DA
% geoyfcc
\newpage
\appendix

\section{Illustrating Cross-domain Robustness}

\begin{figure}
    \centering
    \resizebox{0.48\textwidth}{!}{%
    \includegraphics{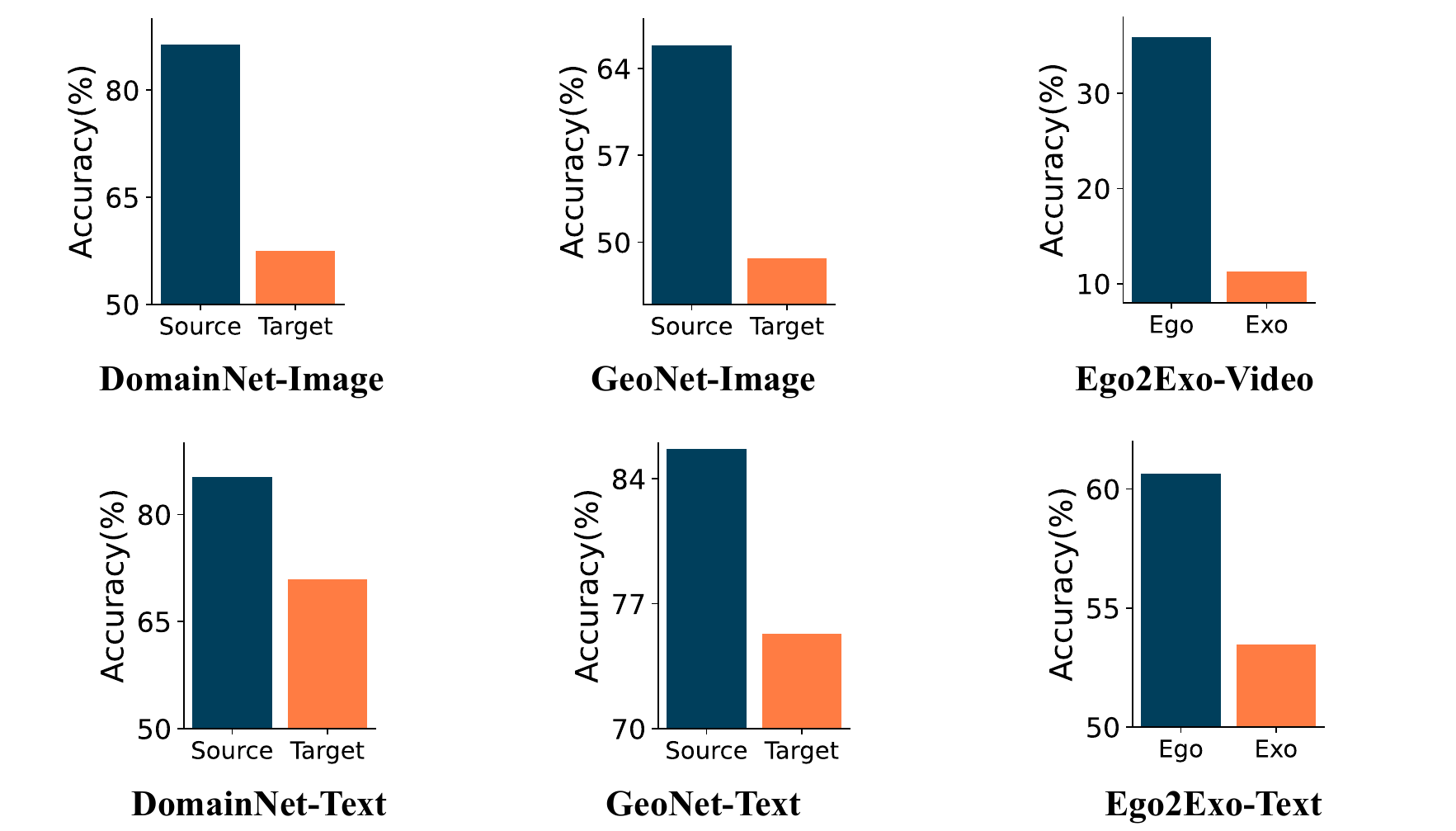}
    }
    \captionsetup{width=0.48\textwidth}
    \caption{{\bf Domain Robustness Text vs Image}: Cross-domain accuracy of image or video classifier transfer across domains compared to text modality. As opposed to significantly large domain drops when transferring image-based models across domains, text-classifier based models are surprisingly robust on all the studied benchmarks, leading to minimal domain drops and high accuracy.
    } 
    \label{fig:supp_pic}
    \vspace{-9pt}
\end{figure}

We illustrate the cross-domain robustness properties of image vs text classifiers in \cref{fig:supp_pic}. We show the remarkably powerful target-domain models obtained by transferring the text classifier, as opposed to image-based models which suffer high domain gap. This behavior is consistent across all the datasets studied, and forms the backbone of our motivations in leveraging textual guidance in performing unsupervised transfer across domains. 

\section{Construction of \egoexo{} benchmark}
\label{sec:egoexo_creation}

We curate \egoexo{} dataset from the larger Ego-Exo4D dataset~\cite{grauman2023ego}.
Specifically, the \textit{keystep annotations} provided along with Ego-Exo4D offer fine-grained action labels for several short video clips, called \textit{segments}, that are manually trimmed from long procedural videos to focus on the keysteps to recognize. We restrict focus to videos from \textit{cooking} activity, as they include the largest set of segments and labels capturing a diverse pool of actions. Moreover, to ease the difficult task of predicting very fine-grained action classes from short segments (eg: \texttt{add coffee beans} vs. \texttt{add coffee grounds}), we use the provided label hierarchy and manually remap the original 96 annotated actions labels into 24 labels by merging similar classes into a larger, common class. 
The final category list is as follows:
\begin{tight_enumerate}
    \item Cook
    \item Serve
    \item Clean up
    \item Add water
    \item Make dough
    \item Make pasta
    \item Make salad
    \item Make chai tea
    \item Make milk tea
    \item Get Ingredients
    \item Prepare dressing
    \item Prepare a skillet
    \item Add spring onions
    \item Turn off the stove
    \item Check paper recipe
    \item Prepare ingredients
    \item Prepare milk (boiled)
    \item Construct undressed salad
    \item Cook noodles in a skillet
    \item Get kitchenware \& utensils
    \item Brew coffee (instant coffee)
    \item Boil noodles in boiling water
    \item Brew coffee (manual pour-over)
    \item Mix noodles with sauce in a bowl
\end{tight_enumerate}
% The action recognition task, therefore, is to predict one of the 22 keystep labels given an action segment as the input. 

To provide text supervision to our algorithm, we use the \textit{atomic action descriptions} provided in Ego-Exo4D dataset. These descriptions provide a narrative of the events in the video, presented in free-form text from the perspective of a third-party observer. Unlike keystep labels, which are defined between specific start and end times within a video, these text descriptions are associated with distinct timestamps, or a single point in time within the video. To create correspondence mapping between the keystep segments and text descriptions, we adopt the method outlined in EgoVLP~\cite{lin2022egocentric} as follows: to generate a text description for a segment, we compile all text descriptions that fall within the timestamps defined by the start and end times of that segment. If multiple timestamps exist, we concatenate the corresponding texts; if no timestamps are available, we include no associated text with the segment. Furthermore, we concatenate the annotations provided by multiple annotators in creating the text description.

Our proposed \egoexo{} consists of video segments labeled with actions from one of the 24 keysteps, with corresponding text descriptions for each segment. We split these video segments into two equal groups classwise, and collect ego-videos from one group and exo-videos from the other to create our adaptation benchmark. The same procedure applied to the validation videos yields 3147 validation segments with both ego and exo views. The \texttt{json} file containing our complete set of videos (referenced from Ego-Exo4D dataset), along with annotations and text descriptions is available along with our publicly released code.

\section{Training Details}
\label{sec:training_details}

\para{Image Classifier} We use a ViT-base~\cite{dosovitskiy2020image} backbone as the image encoder on the GeoNet dataset, and follow prior work~\cite{zhu2023patch} and use Swin-base backbone~\cite{liu2021swin} for experiments on the DomainNet data. Both the backbones are pre-trained on ImageNet-1k, and we add a 2-layer MLP on top of the computed features as the classifier head. 
% For the image classifier backbones, we use a Swin-base backbone~\cite{liu2021swin} for DomainNet dataset following prior works~\cite{zhu2023patch}, and ViT-base~\cite{dosovitskiy2020image} backbone for GeoNet and GeoYFCC datasets, 
% both pre-trained on Imagenet-1k. 
Across all transfer settings, we train these backbones for 90,000 iterations using the objective function specified in \cref{eq:joint_training}, employing SGD with a learning rate of 3e-4 and batch size of 64 from each domain, along with a cosine decay schedule. 

\para{Text Classifier} We use a pre-trained Distill-BERT~\cite{sanh2019distilbert} model from HuggingFace as the sentence classification model $\mathcal{B}(;\phi)$, and fine-tune it for five epochs over the source domain data using AdamW optimizer with a learning rate of 5e-5 and cosine decay over the training schedule. We observed sub-optimal performance using other pre-trained backbones such as T5~\cite{raffel2020exploring}, GPT-2~\cite{radford2019language} or text encoder in CLIP~\cite{radford2021learning} (\cref{subsec:ablation}). 

\para{Video Classifier} We use the pre-computed Omnivore-base~\cite{girdhar2022omnivore} features provided along with the EgoExo4D dataset for training and evaluation. Since different keysteps may be represented by largely different timespans (\cref{fig:segment_duration}), we collect all features that fall within the start and end times of a segment, and pool these features together to form a 1536-dimensional feature representation of that segment. 
We then train a 2-layer MLP classifier on top of these features, using the labeled source feature as well as psuedo-labeled target features following \cref{eq:joint_training}. Note that this training strategy is equivalent to training an MLP classifier on top of frozen Omnivore backbone.
For fair comparison, we follow the same strategy for training all the other baselines as well as prior adaptation methods. For methods that require a temporal sequence of features~\cite{wei2022unsupervised, chen2019temporal}, we sample 8 equally spaced features from the complete set of segment features, and use this feature sequence as input. We follow similar strategy for evaluation, and use features pre-extracted from the validation videos for testing. We use the top-1 accuracy on the validation set for evaluation.

% \section{Extending \Ours}

\end{document}